%% file: main.tex
\documentclass[10pt,twocolumn,letterpaper]{article}

\usepackage{iccv}              


\definecolor{iccvblue}{rgb}{0.21,0.49,0.74}
\usepackage[pagebackref,breaklinks,colorlinks,allcolors=iccvblue]{hyperref}
\usepackage{algorithm,algorithmicx,algpseudocode}
\usepackage{multirow}
\usepackage{graphicx}
\usepackage{wrapfig}
\usepackage{lipsum}
\usepackage{xcolor}
\usepackage[accsupp]{axessibility}
\definecolor{darkgreen}{rgb}{0.0, 0.5, 0.0}

\def\paperID{8542} 
\def\confName{ICCV}
\def\confYear{2025}

\newcommand{\Hquad}{\hspace{0.5em}} 

\title{HazeFlow: Revisit Haze Physical Model as ODE and \\ Non-Homogeneous Haze Generation for Real-World Dehazing}

\author{
Junseong Shin$^{1*}$ \quad
Seungwoo Chung$^{1*}$ \quad
Yunjeong Yang$^{1}$ \quad
Tae Hyun Kim$^{12\dagger}$ \\
$^1$Department of Artificial Intelligence, Hanyang University \\
$^2$Department of Computer Science, Hanyang University \\
{\tt\small \{junsung6140, jsw98, yunjeongyang, taehyunkim\}@hanyang.ac.kr}
}

\begin{document}
\maketitle

\let\thefootnote\relax\footnotetext{$^*$ Equal contribution. \hspace{3pt} $^\dagger$ Corresponding author.
}

\input{sec/0_abstract}    
\input{sec/1_intro}
\input{sec/2_related_works}

\input{sec/3_method}

\input{sec/4_NH.tex}

\input{sec/5_experiments.tex}

\input{sec/6_conclusion.tex}

\input{sec/acknowlege}
{
    \small
    \bibliographystyle{ieeenat_fullname}
    \bibliography{main}
}

\input{sec/X_suppl}

\end{document}

%% file: sec/0_abstract.tex
\vspace{-1em}
\begin{abstract}
\vspace{-2.5 em}

Dehazing involves removing haze or fog from images to restore clarity and improve visibility by estimating atmospheric scattering effects. 
While deep learning methods show promise, the lack of paired real-world training data and the resulting domain gap hinder generalization to real-world scenarios.
In this context, physics-grounded learning becomes crucial; however, traditional methods based on the Atmospheric Scattering Model (ASM) often fall short in handling real-world complexities and diverse haze patterns.
To solve this problem, we propose HazeFlow, a novel ODE-based framework that reformulates ASM as an ordinary differential equation (ODE). 
Inspired by Rectified Flow (RF), HazeFlow learns an optimal ODE trajectory to map hazy images to clean ones, enhancing real-world dehazing performance with only a single inference step. 
Additionally, we introduce a non-homogeneous haze generation method using Markov Chain Brownian Motion (MCBM) to address the scarcity of paired real-world data. 
By simulating realistic haze patterns through MCBM, we enhance the adaptability of HazeFlow to diverse real-world scenarios. 
Through extensive experiments, we demonstrate that HazeFlow achieves state-of-the-art performance across various real-world dehazing benchmark datasets.
Code is available at \url{https://github.com/cloor/HazeFlow}.
\end{abstract}


%% file: sec/1_intro.tex
\vspace{-1.5 em}
\section{Introduction}
\label{sec:intro}
\vspace{-0.5em}

Haze considerably diminishes visual quality, thereby posing challenges for diverse real-world applications, including autonomous driving, aerial surveillance, and the analysis of outdoor scenes~\cite{huang2020dsnet, li2023domain}.
The non-homogeneous, spatially varying nature of real-world haze makes dehazing even more complex.
Despite extensive research, there remains a substantial performance gap between synthetic and real-world scenarios.

\begin{figure}[t!]
    \centering
    \includegraphics[width=\linewidth]{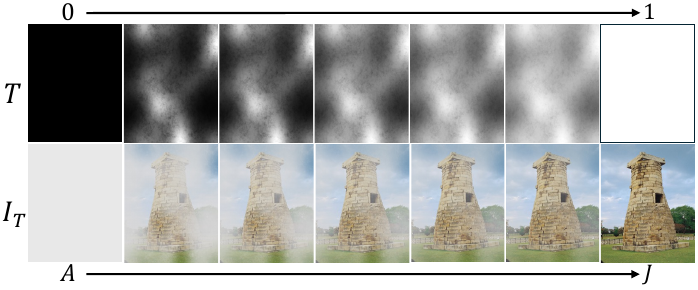}
    \vspace{-2.0 em}
    \caption{Illustration of dehazing trajectory of hazy image $I_T$ as transmission map $T$ gradually increases from 0 to 1.}
    \vspace{-1.5 em}
    \label{fig:ODE}
\end{figure}

Hazy images are typically represented by the Atmospheric Scattering Model (ASM)~\cite{narasimhan2003asm1, narasimhan2002asm2}, which captures the interaction of light with haze particles as it travels through the atmosphere.
ASM is formulated as:
\vspace{-0.5em}
\begin{equation}
\label{eq:1}
    I(x) = T(x)J(x) + (1-T(x))A,
\end{equation}
where $x$ represents a pixel location, $I$  denotes the observed hazy image, $J$ is the clear scene radiance, $A$ is the global atmospheric light, and $T$ is the transmission map.
The transmission map quantifies the degree of haze at the pixel location $x$, and is defined as:
\vspace{-0.5em}
\begin{equation}
    \label{eq:transmission}
    T(x) = e^{-\beta d(x)},
\end{equation}
where $\beta$ denotes the haze density coefficient and $d(x)$ represents the scene depth.
For example, as shown in Fig.~\ref{fig:ODE}, $T = 0$ results in complete opacity with $I$ converge to $A$, while $T = 1$ reveals a clear scene with $I$ matches $J$.

Traditional physical-based haze removal methods~\cite{he2010single, zhu2014single, fattal2014dehazing}, often rely on ASM, provide a principled understanding of how haze affects images, but physical models alone struggle to capture the complexity of real-world haze.
On the other hand, deep learning approaches~\cite{cai2016dehazenet, li2017aod, qin2020ffa, guo2022dehamer, im2023deep} trained on large-scale synthetic datasets~\cite{li2018benchmarking} achieve impressive results on curated benchmarks, but fall short in real-world settings due to domain gap. 
Techniques such as data augmentation-based approaches~\cite{chang2024panet}, CycleGAN-based methods~\cite{zhu2017unpaired, chen2022unpaired}, and recent diffusion-based models~\cite{luo2023daclip, zheng2024diffuir} have indeed improved robustness. However, handling real-world haze artifacts and distribution shifts remains a major challenge.
To fully leverage both physical models and deep learning methodologies, we introduce HazeFlow, a novel ODE-based dehazing framework that redefines dehazing as an optimal transport process between haze and clean image distributions.
Instead of directly solving an inverse problem of ASM, we reformulate dehazing as an ordinary differential equation (ODE), where the velocity field guides hazy images along a physically grounded path to their clean counterparts.
This reformulation enables the utilization of contemporary neural ODE methodologies to facilitate the training of a dehazing network across the complete dehazing trajectory, thereby enhancing the stability, accuracy, and efficiency of haze removal.
Motivated by Rectified Flow (RF)~\cite{liu2022flow} which learns straight, fast sampling trajectories, we train a neural network to approximate the optimal ODE trajectory, allowing fast and accurate removal of the haze in a single inference step.

Furthermore, HazeFlow enhances real-world performance with a three-stage learning process: Pretrain, Reflow, and Distillation.
Initially, the Pretrain phase on large-scale synthesized hazy-clean image pairs constructs a mapping from hazy images to their clean counterparts.
The Reflow stage fine-tunes the model using real hazy and pseudo clean image pairs, capturing complex real-world haze patterns.
Finally, the Distillation stage further improves robustness to real-world artifacts and enhances perceptual quality.
Through three-stage learning, HazeFlow develops a dynamic, transmission-aware mapping, enhancing its understanding of real-world haze.
HazeFlow demonstrates superior adaptability to real-world scenarios, achieving state-of-the-art results in both synthetic and real-world benchmark datasets.

In addition, we tackle the issue of the limited availability of paired real-world datasets.
Datasets containing non-homogeneous haze (e.g., smoke-induced haze) are scarce, and most synthetic datasets assume homogeneous haze with a constant scattering coefficient $\beta$ across the scene~\cite{li2018benchmarking, sakaridis2018semantic, wu2023ridcp}.
To mitigate this limitation, we introduce a non-homogeneous haze generation method that utilizes Markov Chain Brownian Motion (MCBM) which effectively models the random movements of airborne particles~\cite{van1992stochastic, ito2012diffusion}.
Specifically, we simulate a spatially varying scattering coefficient for transmission map using MCBM.
This provides a more realistic and diverse dataset, allowing the model to generalize better to real-world haze.

The contributions of this paper are as follows.
\begin{itemize}
    \item For the first time, we reformulate the dehazing process as solving the ODE of ASM, enabling the learning of a physics-guided flow.
    Our three-stage training pipeline enhances real-world generalization and achieves stable one-step estimation.
    \item We propose a non-homogeneous haze image generation technique with MCBM modeling to mitigate the limited availability of paired real-world hazy datasets.
    \item Our method achieves state-of-the-art performance in real-world dehazing tasks across various benchmark datasets, demonstrating exceptional efficacy in practical applications.
\end{itemize}

%% file: sec/2_related_works.tex
\vspace{-0.5em}
\section{Related Works}
\label{sec:related_works}

\vspace{-0.25em}
\subsection{Single Image Dehazing}
\vspace{-0.5em}
Early methods relied on assumptions or priors to estimate the haze.
DCP~\cite{he2010single} assumes that haze-free images have at least one low intensity channel.
Other priors, such as the color attenuation prior~\cite{zhu2014single}, color-lines~\cite{fattal2014dehazing}, Haze-lines~\cite{berman2018single} used color and intensity gradients to predict transmission map.
However, these methods often struggle with complex scenes, sky regions, and non-uniform haze, producing undesired artifacts. 


Recently, deep learning approaches have significantly improved dehazing performance by directly learning to remove haze from data.
Early methods relied heavily on Convolutional Neural Networks (CNNs), which are effective in capturing local characteristics and hierarchical information~\cite{ren2016single, cai2016dehazenet,li2017aod, liu2019griddehazenet, dong2020multi, wu2021contrastive,qin2020ffa}.
Recent research has explored Transformer-based architectures, which excel in modeling global relationships and capturing long-range dependencies within the image~\cite{guo2022dehamer, song2023vision}.
However, these approaches face challenges when applied to real-world hazy images, as they often overlook the underlying physics model of haze formation.
Without explicitly modeling the atmospheric scattering process and spatially varying transmission, these methods can struggle to generalize across diverse haze patterns, leading to incomplete dehazing or distortion of fine details.

\vspace{-0.25em}
\subsection{Real-World Dehazing}
\vspace{-0.5em}
\begin{figure*}[th!]
    \centering
    \includegraphics[width=1.0\linewidth]{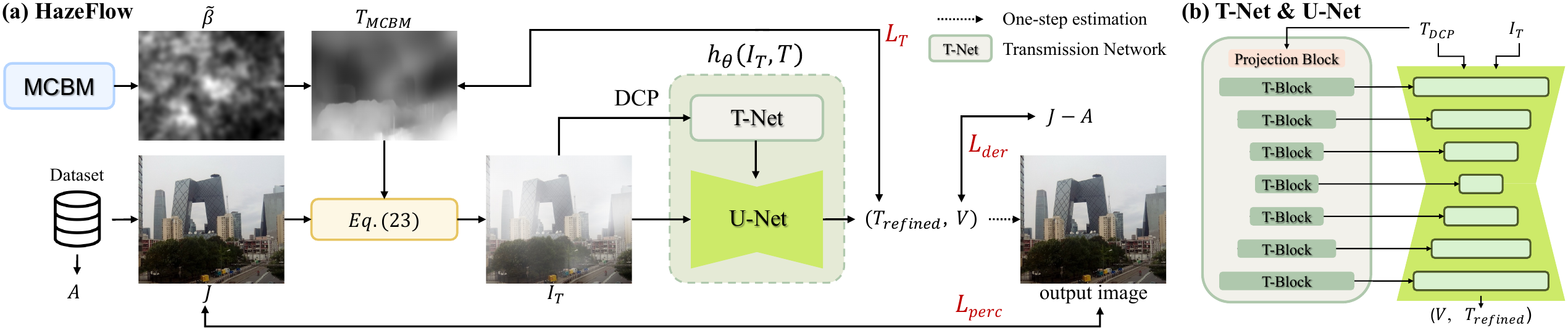}
    \vspace{-3.0ex}
    \caption{Overview of our proposed method, illustrating (a)~HazeFlow and (b)~Transmission Network.}
    \label{fig:process}
    \vspace{-3.5ex}
\end{figure*}

To mitigate the domain gap between synthetic and real-world hazy images, various approaches have been proposed.
Methods based on CycleGAN~\cite{chen2022unpaired, yang2022self, shao2020dad, li2022physically} utilize cycle consistency loss~\cite{zhu2017unpaired} to tailor models for real-world scenarios by employing unpaired datasets of hazy and clean images.
On the other hand, RIDCP~\cite{wu2023ridcp} tackled the scarcity of real-world haze data by generating a dataset with synthesized haze applied to hundreds of clean images.
Despite these efforts, challenges persist in effectively modeling non-homogeneous real-world haze.
PANet~\cite{chang2024panet} attempts to address this issue by augmenting non-homogeneous haze.
Specifically, it modifies the values of $\beta$ or A to augment real-world haze.
However, this method is limited to augmentations within limited datasets. 
Furthermore, it struggles to fully encompass the diverse haze conditions encountered in real-world scenarios such as in RTTS~\cite{li2018benchmarking}.
In this work, we aim to address these limitations by providing an enhanced dataset with more diverse haze characteristics, enabling more reliable training and generalization.

\vspace{-1em}
\paragraph{Diffusion-based Image Restoration}
Recently, diffusion models have shown remarkable success in image restoration tasks by iteratively refining images through learned denoising processes.
To address real-world dehazing, DehazeDDPM~\cite{yu2023high} and Wang~\textit{et al.}~\cite{wang2024frequency} leverage the diffusion model~\cite{ho2020denoising} to improve dehazing performance by applying conditional diffusion models along with additional refinement techniques.
DACLIP~\cite{luo2023daclip} and DiffUIR~\cite{zheng2024diffuir} proposed unified frameworks capable of handling multiple restoration tasks, including dehazing through a shared diffusion process.
Despite their impressive results, these diffusion-based methods face limitations in inference speed and computational efficiency. 
In contrast, our methodology employs a flow-based strategy that uses a well-defined ODE formulation, significantly decreasing the number of sampling steps required during the inference process while preserving superior restoration quality.

%% file: sec/3_method.tex
\vspace{-0.5em}
\section{ODE-based Dehazing Model}
\label{sec:method}
\vspace{-0.5em}
This section begins with a concise summary of RF, which serves as an inspiration for our study, and is followed by an in-depth description of our proposed method, HazeFlow.


\subsection{Preliminary}
\label{sec:4.1}
RF solves the optimal transport problem between two data distributions $\pi_0$ and $\pi_1$ by transporting data from an initial state $X_0 \sim \pi_0$ to a target state $X_1 \sim \pi_1$ over a continuous time interval $t \in [0, 1]$.
Given observations of $X_0$ and $X_1$, RF learns the velocity field by training a neural network $r_\theta$ to follow the linear interpolation of $X_t = tX_1 + (1-t)X_0$.
This is achieved by solving the following least squares regression problem:
\vspace{-0.5em}
\begin{equation}
    L(\theta) = \underset{\theta}{min} \int_0^1 \mathop{\mathbb{E}}[||(X_1 - X_0) - r_\theta(X_t,t)||^2] \,\mathrm{d}t.
\end{equation}
After training the velocity function $r_\theta$, target $X_1$ can be inferred by a conventional ODE-solver (\textit{e.g.} Euler solver):
\vspace{-0.5em}
\begin{equation}
\small
    X_{k+\frac{1}{N}} = X_k + \frac{1}{N}r_{\theta}(X_k, k), \Hquad \text{s.t.} \Hquad  k \in \left\{0, \frac{1}{N}, \cdots, \frac{N-1}{N} \right\},
\end{equation}
where $N$ denotes number of steps from $X_0$ to $X_1$, and $k$ is the discrete step index.

\vspace{1ex}
\noindent\textbf{Reflow.}
To straighten the trajectory and yield fast simulation, RF introduces Reflow mechanism.
Using $r_\theta$ trained on pairs of ($X_1$, $X_0$), new synthetic pairs ($\hat{X_1}$, $X_0$) can be generated, where $\hat{X_1}$ is obtained from $X_0$ using a conventional ODE solver (\ie, $ODE[r_\theta](X_0))$).
The newly generated pair is used for training to refine the velocity field through $r_\phi$.
This process is repeated iteratively to streamline the path.
The aim of the objective function is to ensure that the predictions of $r_\phi$ correspond with the velocity field from $X_0$ to $\hat{X_1}$, resulting in more direct trajectories, as follows:
\vspace{-0.5em}
\begin{equation}
    \label{eq:RF_reflow}
    L(\phi) = \underset{\phi}{min} \int_0^1 \mathop{\mathbb{E}}[||(\hat{X_1} - X_0) - r_\phi(X_t,t)||^2] \,\mathrm{d}t, 
\end{equation}
where $X_t = t\hat{X_1} + (1-t)X_0$. 
By refining the path using data pairs derived from solving the ODE, the newly trained $r_\phi$ achieves accurate results with fewer inference steps.

\vspace{1ex}
\noindent\textbf{Distillation.}
Using $r_\phi$ from Reflow, a network $r_{\phi'}$ requiring one-step inference can be obtained via knowledge distillation, as follows:
\vspace{-0.5em}
\begin{equation}
    \label{eq:RF_distill}
    L(\phi') = \mathbb{E}[\mathbb{D}(ODE[r_\phi](X_0),\ X_0 + r_{\phi'}(X_0, 0))],
\end{equation}
where the loss function $\mathbb{D}$ uses LPIPS~\cite{zhang2018lpips}, which enhances the perceptual quality of images produced.
Through these Reflow and Distillation processes, RF enables high-quality one-step generation and has been successfully applied in various tasks~\cite{zhu2024flowie, wang2024semflow, liu2023instaflow}.



\subsection{HazeFlow}
\label{sec:4.2}

\noindent\textbf{Revisit ASM as an ODE.}
In this work, our key observation is that the conventional ASM in Eq.~\ref{eq:1} can be reformulated as an ODE: 
\vspace{-0.5em}
\begin{equation}
    \label{eq:ODE}
    \mathrm{d}I_T = (J - A) \mathrm{d}T,
\end{equation}
where $I_T$ is the hazy image corresponding to the transmission map $T$.
This reformulation reveals that dehazing fundamentally involves modeling incremental changes in $T$ and proportional adjustments in $I_T$.
In this view, the dehazing process becomes a velocity estimation problem, where the goal is to learn a vector field that guides the evolution of the hazy image towards the clean image along the $T$ path.
We achieve this by parameterizing the derivative function using RF~\cite{liu2022flow} through a learnable neural network $h_\theta$, allowing the model to approximate the optimal transport trajectory from atmospheric light to clean image as in Fig.\ref{fig:ODE}.

Using the RF framework, our method produces diverse and high-quality results and drastically reduces computational costs compared to conventional iterative models (\eg, diffusion models).
Once $h_\theta$ is trained, clean image $J$ can be directly predicted from the hazy image $I_{T}$ with an off-the-shelf Euler ODE-solver as follows:
\vspace{-0.5em}
\begin{equation}
    \begin{aligned}
        I_{\tau+ \frac{1}{M}} &= I_\tau + \frac{1}{M} h_\theta(I_\tau,\tau),  \\
        &\text{s.t.} \quad \tau \in \left\{T, T+\frac{1}{M}, \ldots, T+\frac{N-1}{M} \right\},
    \end{aligned}
    \label{eq:ode_solver}
\end{equation}
where $\frac{1}{M}$ denotes the adaptive step size, defined as $\frac{1-T}{N}$, where $N$ is the number of sampling steps from $I_T$ to $I_1(=J)$.
Unlike conventional RF-based models~\cite{zhu2024flowie, wang2024semflow, liu2023instaflow}, which apply a time variable $t$ uniformly across the entire image, our method assigns a locally varying transmission map $T$ across images.
Since the degree of attenuation varies for each pixel in a hazy image, the transmission map $T$ exhibits spatial variability in the image.
This spatial variation is managed through the adaptive step size: areas with thicker haze (\ie, lower $T(x)$) undergo more transformations, whereas regions with lighter haze (\ie, higher $T(x)$) require fewer updates, thereby preserving finer details.
\subsection{Network Architecture}
Our HazeFlow consists of a U-Net and a T-Net, as shown in Fig.~\ref{fig:process} (b).
The U-Net takes the hazy image and transmission map derived from the hazy image as inputs, while the T-Net processes the transmission map.
Input transmission map $T_{DCP}$ is estimated using DCP~\cite{he2010single}.
T-Net then embeds the transmission feature, which is combined with the U-Net to predict the dehazed image.

\vspace{1ex}
\noindent\textbf{Transmission Network (T-Net).}
We propose T-Net to effectively handle the spatially varying input transmission map $T$ in a high-dimensional feature space as shown in Fig.~\ref{fig:process} (b).
To achieve this, input $T_{DCP}$ is projected into the feature space by a Projection Block comprising two $1\times1$ convolution layers, which generates the representation in the feature space $T^{l=0}$.
Subsequently, the Transmission Block (T-Block), which consists of two convolution layers and SiLU~\cite{elfwing2018silu} activation functions, operates at multiple scales $l\in \{0,1,2,3\}$.
It takes $T^l$ as input and generates $T^{l+1}$ at scale level $(l+1)$ as:
\vspace{-0.5em}
\begin{equation}
    T^{l+1} = Conv_{1\times1}(SiLU(Conv_{3\times3}(SiLU(T^l)))).
    \label{eq:transmission_block.}
\end{equation}
The embedding $T^l$ is integrated with U-Net features to effectively handle spatially varying transmissions.
Each $T^l$ is utilized in both the encoder and decoder of the U-Net.

\vspace{1ex}
\noindent\textbf{Transmission Refinement (T-Refinement).}
The transmission map $T_{DCP}$ derived from DCP can be erroneous in real-world hazy conditions, especially in regions like sky where its underlying assumptions break down.
To address this problem, we refine $T_{DCP}$ in our HazeFlow by jointly predicting velocity and improved transmission map as follows:
\vspace{-0.5em}
\begin{equation}
    (V, T_{refined}) = h_\theta(I_T, T_{dcp}),
\end{equation}
where $V$ denotes velocity and $T_{refined}$ represents the improved transmission map.
Specifically, we employ the ground-truth transmission map $T_{gt}$ to train the network and obtain the refined transmission map by minimizing the loss as: 
\begin{equation}
    L_{T}(\theta) = \mathop{\mathbb{E}}||T_{gt} - T_{refined}||^2_2.
\end{equation}
This approach ensures that the refinement aligns closely with the realistic haze characteristics defined by $T_{gt}$.
During the inference phase, the refined transmission map is used as initial $\tau$ in Eq.~\ref{eq:ode_solver}, resulting in more accurate and visually coherent haze removal.
Refer to the Supplementary Material Sec.~\ref{sec:supp_transmission refinement} for additional details.

In the remainder of the study, when the output of $h_\theta$ is not explicitly stated, it refers to $V$ for simplicity.

\subsection{Three-Stage Learning}
We introduce a three-stage learning scheme including Pretrain, Reflow, and Distillation. Through this approach, we progressively adapt the model to real-world scenarios, enabling real-world dehazing with fewer sampling steps.

\vspace{1ex}
\noindent\textbf{Pretrain}
For the Pretrain phase, we first introduce a loss function $L_{der}$ to train the neural network $h_\theta$ on large-scale synthesized hazy-clean pairs, learning the derivative function:
\vspace{-0.5em}
\begin{equation}
    \label{eq:haze_loss_der}
    \small
    L_{der}(\theta) = \underset{\theta}{\min} \int_0^1 \mathop{\mathbb{E}} \left[ \left\| (J - A) - h_\theta(I_T, T) \right\|^2 \right] \,\mathrm{d}T.
\end{equation}
Moreover, to enhance perceptual quality of the dehazed image, we incorporate the LPIPS-based loss $L_{perc}$ using a one-step estimation approach as:
\vspace{-0.5em}
\begin{equation}
    \label{eq:lpips}
    L_{perc}(\theta) = \mathbb{E}[\mathbb{D}(J,\ I_T + (1-T)h_\theta(I_T, T))].
\end{equation} 
We observe that this accelerates training by yielding perceptually satisfactory images~\cite{lee2024improving}.
Finally, by combining the derivative, perceptual, and transmission refinement losses, the total loss function for training $h_\theta$ in the Pretrain stage is defined as:
\vspace{-0.5em}
\begin{equation}
    L(\theta) = L_{der}(\theta) + L_{perc}(\theta) + w *L_{T}(\theta),
\end{equation}
where $w$ is a users-defined parameter to control the weight of the transmission refinement loss $L_{T}(\theta)$.

\vspace{1ex}
\noindent\textbf{Reflow.}
The Reflow stage trains a new RF model $h_\phi$ (sharing the same architecture as $h_\theta$) to adapt to realistic haze distributions.
In particular, we tackle two major challenges: (1) the lack of clean ground-truth counterparts corresponding to real-world hazy images, and (2) the complexity of accurately modeling the diverse distributions of atmospheric light $A$~\cite{chang2024panet}.
Specifically, when a real-world hazy image $I^{reael}_T$ without corresponding ground truth counterpart is given, the pretrained $h_\theta$ can generate a pseudo clean image $\hat{J}$ and spatially-varying atmospheric light $\hat{A}$ through one step estimation as follows:
\begin{align}
    &\hat{J} = I^{real}_T + (1-T)\cdot h_\theta(I^{real}_T, T),  \\
    &\hat{A} = I^{real}_T - T\cdot h_\theta(I^{real}_T, T).
\end{align} 
While $h_\theta$ from the Pretrain stage assumes a uniform atmospheric light $A \in \mathbb{R}^3$ over the entire image, Reflow relaxes this assumption by estimating a spatially varying $\hat{A}$ in $\mathbb{R}^{H\times W \times 3}$ from real-world images. 
This allows the model to adapt to a much broader spectrum of real-world haze distributions.

Then the pseudo pair ($\hat{J}$, $\hat{A}$) serves as a proxy for ground truth, enabling further training of $h_\phi$ with the loss given by,
\vspace{-0.5em}
\begin{equation}
    \label{eq:our_reflow}
    L^{real}_{der}(\phi) = \underset{\phi}{min} \int_0^1 \mathop{\mathbb{E}}[||(\hat{J} - \hat{A}) - h_\phi(\gamma(I^{real}_T),T)||^2] \,\mathrm{d}T, 
\end{equation}
where $\gamma$ represents random gamma correction~\cite{wu2023ridcp} to predict appropriate brightness.
To further enhance perceptual quality, we incorporate the LPIPS loss, yielding the final objective to train $h_\phi$ as:  
\vspace{-0.25em}
\begin{equation}
    L_{Reflow}(\phi) = L^{real}_{der}(\phi) + L_{perc}(\phi). 
\end{equation}
Notably, acquiring a ground truth transmission map for the real-world $I^{real}_T$ is impractical, hence transmission refinement is not performed at this stage.

\vspace{1ex}
\noindent\textbf{Distillation.}
In Distillation stage, we further improve the robustness of the model against real-world artifacts by distilling $h_\phi$ to a new model $h_\phi'$ (with the same architecture as $h_\phi$).
Specifically, we improve performance by suppressing unwanted degradations and enhancing output quality.
To achieve this, the LPIPS loss is employed, as follows:
\begin{equation}
\begin{aligned}    
    L_{Distill}(\phi') = 
    &\mathbb{E} \big[ \mathbb{D}(I_T + (1 - T) h_\phi(I_T, T), \\
    & \mathcal{G}(I_T) + (1 - T) h_{\phi'}(\mathcal{G}(I_T), T)) \big],
    \label{eq:our_distill}
\end{aligned}
\end{equation}
where $\mathcal{G}$ represents real-world data augmentation, consisting of various degradations~\cite{fang2024real}. 
This augmentation forces $h_\phi’$ to learn how to remove various real-world artifacts, making the dehazing process more robust to unseen degradation patterns. 
Moreover, the Distillation can be iterated multiple times to progressively improve performance, leading to cleaner and more visually appealing results.

%% file: sec/4_NH.tex

\vspace{-0.5em}
\section{Non-Homogeneous Haze Generation}
\label{sec:NH}
\noindent\textbf{Non-Homogeneous Haze.}
In ASM, the thickness of the haze is determined by the value of the transmission map $T(x)$, which depends on scene depth $d(x)$ and the haze density coefficient $\beta$. 
Previous methods generate haze by sampling a scalar $\beta$ for the entire image~\cite{wu2023ridcp, li2018benchmarking}, which fails to capture real-world non-homogeneous characteristics, such as smoke~\cite{ancuti2018ohaze}.
Vinay et al.~\cite{shetty2023non} aimed to create realistic haze using Photoshop; however, producing various non-homogeneous hazy images demands substantial resources.
To address this issue, we draw inspiration from the concept that air particle movement can be modeled using Brownian motion~\cite{ito2012diffusion, van1992stochastic} and utilize Markov Chain Brownian Motion (MCBM) to generate haze with non-homogeneous characteristics by spatially varying $\beta$ and $d(x)$.

\begin{figure}[t]
    \centering
    \includegraphics[width=\linewidth]{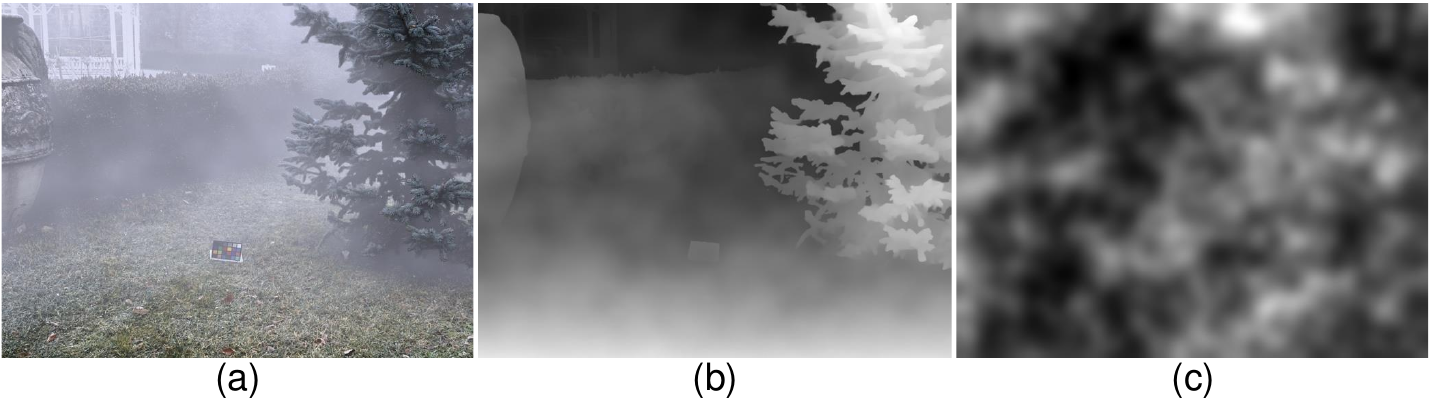}
    \caption{Example of non-homogeneous haze synthesized via MCBM. (a) Generated hazy image. (b) Transmission map $T_{MCBM}$. (c) Spatially varying density coefficient map $\Tilde{\beta}$.}
    \vspace{-1.5em}
    \label{fig:NH_examples}
\end{figure}

\vspace{1ex}
\noindent\textbf{Haze Synthesis using MCBM.}
To simulate non-homogeneous real-world haze, we employ MCBM to generate a spatially varying haze density map $\Tilde{\beta}$.
First, a 2D array is initialized to represent the haze density map on an image grid. 
We begin with a randomly selected pixel location $(i, j)$ within the haze density map. 
This initial coordinate $(i,j)$ is subsequently updated as follows:
\vspace{-0.5em}
\begin{equation}
     (i', j') = (i+a, j+b),
     \label{eq:nh_1}
\end{equation}
where $(a,b) \in \{(0,1),(0,-1),(1,0),(-1,0)\}$ are selected randomly to update the coordinates.
To enhance the non-uniformity of the haze density, we employ the Brownian motion.
Then, the non-uniform haze density map $\Tilde{\beta}$ is updated accordingly as follows:
\vspace{-0.5em}
\begin{equation}
    \Tilde{\beta}(i'+ \Delta i, j'+ \Delta j) =\Tilde{\beta}(i'+ \Delta i, j'+ \Delta j) + 1.
    \label{eq:nh_2}
\end{equation}
where $\Delta i, \Delta j \sim \mathcal{N}(0, \sigma^2)$ are the random displacements for each pixel from Brownian motion, rounded to the nearest integer, with $\sigma$ representing the standard deviation.
In this way, we update the haze density map $\Tilde{\beta}$ progressively for $n$ iterations where $n$ denotes the number of updates in Eq.~\ref{eq:nh_1}.
This process models the random movement of haze particles, emphasizing their non-homogeneous characteristics.
Finally, the haze density map, $\Tilde{\beta}$, is smoothed with a Gaussian filter to produce a more realistic haze density map and then normalized to the range of 0 to 1.
Additional details about MCBM are provided in Supplementary Material Sec.~\ref{supp:mcbm}.

\begin{table*}[hbt!]
    \centering\resizebox{0.85\textwidth}{!}{
  \begin{tabular}{l|cccc|cccc}
    \toprule
    \cmidrule(r){2-5} \cmidrule(r){6-9}
     \multirow{2}{*}{Method}& \multicolumn{4}{c}{RTTS~\cite{li2018benchmarking}} & \multicolumn{4}{c}{Fattal's~\cite{fattal2008single}}\\ 
    \cmidrule(r){2-5} \cmidrule(r){6-9}
    & FADE$\downarrow$ & BRISQUE$\downarrow$&NIMA$\uparrow$  & PAQ2PIQ$\uparrow$ & FADE$\downarrow$ & BRISQUE$\downarrow$ &NIMA$\uparrow$  & PAQ2PIQ$\uparrow$ \\ 
    \midrule
    Hazy input                  & 2.484& 36.64& 4.48 & 66.05& 1.061& 21.08& 5.38 & 71.54        \\
    PDN~\cite{yang2018proximal} & 0.876 & 30.81 & 4.46 & - & -& -& - & -                        \\
    MBSDN~\cite{dong2020multi}    & 1.363 & 27.67 & 4.53 & 66.85 & 0.579 & 14.15 & 5.43 & 72.84 \\
    Dehamer~\cite{guo2022dehamer}  & 1.895 & 33.24 & 4.52 & 66.70 & 0.698 & 15.53 & 5.16 & 71.43\\
    \midrule
    DAD~\cite{shao2020dad}      & 1.130& 32.24& 4.31& 66.79& 0.484& 29.64& \textbf{5.46}& 71.56 \\ 
    PSD~\cite{chen2021psd}      & 0.920 & 27.71 & 4.60 & 70.43& 0.416& 23.61& 4.99& 76.02       \\
    D4~\cite{yang2022self}      & 1.358 & 33.21 & 4.48 & 66.84& 0.411& 20.33& 5.44 & 73.13      \\
    RIDCP~\cite{wu2023ridcp}    & 0.944 & 17.29 & 4.97 & 70.82& 0.408& 20.05& 5.43 & 74.64      \\
    CORUN~\cite{fang2024real}   &0.824 & 11.96	& \textbf{5.34} &72.56	&0.338&14.82&5.39&76.12 \\
    \midrule
    \midrule
    HazeFlow &\textbf{0.583} &\textbf{5.01}&	5.30&\textbf{72.97} &\textbf{0.264}&\textbf{13.36}&5.40&\textbf{76.44}\\ 
    \bottomrule	
    \end{tabular}}
    \vspace{-1.0ex}
\caption{Quantitative results on unpaired dataset (RTTS~\cite{li2018benchmarking}, Fattal~\cite{fattal2008single}). 
Best results are \textbf{bolded}.}
\vspace{-1.0ex}
\label{tab:quantitativeunpaired}
\end{table*}

\begin{figure*}[ht]
    \centering
    \includegraphics[width=\linewidth]{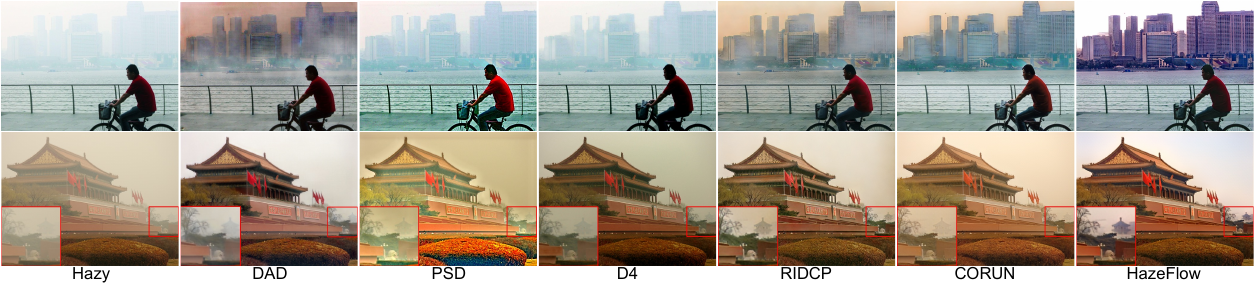}
    \vspace{-4Ex}
    \caption{Qualitative unpaired dataset comparison on RTTS~\cite{li2018benchmarking}~(top row) and Fattal~\cite{fattal2008single}~(bottom row).}
    \label{fig:unpaired}
    \vspace{-3.5Ex}
\end{figure*}

\vspace{1ex}
\noindent\textbf{Haze Dataset Generation.}
Based on MCBM, we generate non-homogeneous transmission map by combining the uniform $\beta$ and non-uniform $\Tilde{\beta}$ as:
\vspace{-0.5em}
\begin{equation}
\label{eq:T_MCBM}
    {T}_{MCBM} = e^{-(\beta + \alpha \cdot \Tilde{\beta})d},
\end{equation}
where the random parameter $\alpha \in [0.5,1]$ controls the degree of non-homogeneity.
Subsequently, we generate non-homogeneous haze image using ASM: 
\vspace{-0.5em}
\begin{equation}
    {I} = \mathcal{D}({T_{MCBM}}\cdot J  + (1-T_{MCBM})\cdot A),
\end{equation}
where $\mathcal{D}$ represents the real-world degradation process, such as as Gamma correction, additive Gaussian noise, and JPEG compression, following~\cite{wu2023ridcp}.


In practice, to generate a large number of non-homogeneous haze images, we use RIDCP500~\cite{wu2023ridcp} dataset for the clean image $J$, and RA-depth~\cite{he2022ra} is used to estimate the corresponding scene depth $d$.
The global atmospheric light $A$ is uniformly sampled in the range $[0.25,1.8]$ accounting for the color bias of atmosphere light, and $\beta$ is uniformly sampled from the range $[0.2,2.8]$.
In Fig.~\ref{fig:NH_examples}, (a) shows an example of our synthetic hazy image, (b) illustrates the \( T_{MCBM} \), and (c) represents the spatially varying \( \Tilde{\beta} \) generated by MCBM.
This process produces realistic non-homogeneous haze effects, and the resulting dataset is used during the Pretrain stage of HazeFlow.


%% file: sec/5_experiments.tex
\vspace{-0.5em}
\section{Experiments}

\begin{figure*}[t]
    \centering
    \includegraphics[width=\linewidth]{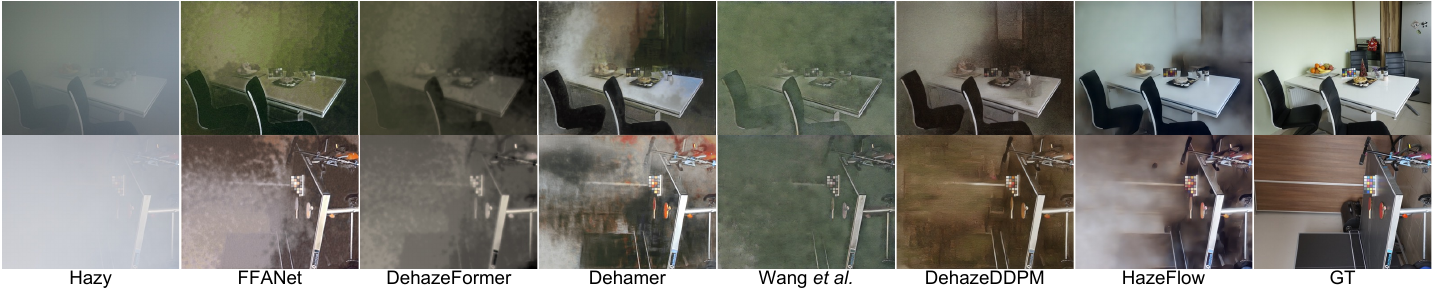}
    \vspace{-4.5 Ex}
    \caption{Qualitative paired dataset comparison on Dense-HAZE~\cite{ancuti2019dense}.}
    \label{fig:paired}
    \vspace{-3.5 Ex}
\end{figure*}

\subsection{Experimental Setting}
\noindent\textbf{Datasets.}
In the Pretrain stage, we use a synthetic non-homogeneous dataset generated by our MCBM method with the RIDCP500~\cite{wu2023ridcp}, consisting of 500 clear images paired with depth maps estimated using~\cite{he2022ra}.
In experiments utilizing unpaired datasets ($I^{real}_{T}$ without the corresponding ground truth $J$), we use the URHI dataset~\cite{li2018benchmarking}, which comprises 4,807 real hazy images, for Reflow and Distillation stages.
In experiments utilizing paired datasets of clean and hazy images, we use paired training datasets ($I^{real}_{T}$ with the corresponding ground truth $J$) in the Reflow and Distillation phases to maintain a fair comparison with alternative approaches.
Additional experiments in which the model is trained on the URHI dataset instead of the training data from the paired dataset can be found in the Supplementary Material Sec.~\ref{sup:paired}.

To evaluate real-world dehazing performance, we use unpaired real-world haze datasets, including RTTS~\cite{li2018benchmarking}, which consists of 4,322 hazy images, and Fattal’s dataset~\cite{fattal2008single} containing 31 classic hazy images.
Moreover, for paired real-world haze datasets, we use the NH-HAZE~\cite{ancuti2020nh} and Dense-Haze~\cite{ancuti2019dense} datasets, both of which contain 50 training data pairs and 5 test data pairs.

\vspace{1ex}
\noindent\textbf{Implementation Details.}
The baseline U-Net architecture of HazeFlow follows NCSN++~\cite{songscore}.
In the Pretrain stage, we train HazeFlow for 100K iterations using hazy images generated by our MCBM from the RIDCP500 dataset.
We use Adam~\cite{kingma2014adam} optimizer and Cosine annealing~\cite{loshchilov2016sgdr} with an initial learning rate of $2 \times 10^{-4}$ to $1 \times 10^{-6}$ for optimization. 
The weight of the transmission refinement loss is set to $w = 0.5$.
To focus on learning the derivative function of the ODE, we fix $A$ to $[1,1,1]$.

Next, in the Reflow and Distillation phases, the network is trained for only 10K iterations with Adam~\cite{kingma2014adam} optimizer.

For inference, we perform single-step generation with $N=1$.

\vspace{1ex}
\noindent\textbf{Evaluation Metrics.}
For evaluation on unpaired datasets, we use the Fog-Aware Density Evaluator (FADE)~\cite{choi2015referenceless} to measure the haze density. 
Furthermore, to evaluate the natural appearance and ensure our results lack any artifacts, we utilize BRISQUE~\cite{mittal2012brisque}. Meanwhile, NIMA~\cite{talebi2018nima} is employed for a thorough assessment of both image quality and aesthetics.
Additionally, we incorporate PAQ2PIQ~\cite{ying2020paq2piq} for a further perceptual quality comparison.
For evaluation on paired datasets, we measure the Peak Signal-to-Noise Ratio (PSNR), Structural Similarity (SSIM) and Learned perceptual image patch similarity (LPIPS) in the RGB color space.

\vspace{-0.5 em}
\subsection{Comparative Evaluation}

\vspace{-0.5 em}
\noindent\textbf{Quantitative Results.}
First, for evaluation on the unpaired real-world RTTS and Fattal's datasets,
we compare our method with the SOTA approaches, including PDN~\cite{yang2018proximal}, MSBDN~\cite{dong2020multi} and Dehamer~\cite{guo2022dehamer}, and real-world dehazing methods such as DAD~\cite{shao2020dad}, PSD~\cite{chen2021psd}, D4~\cite{yang2022self}, RIDCP~\cite{wu2023ridcp}, and CORUN~\cite{fang2024real}.
In Tab.~\ref{tab:quantitativeunpaired}, our approach surpasses other methods in dehazing performance across the majority of evaluation metrics for both datasets.
In particular, on RTTS, our method surpasses the second-leading approach by a significant margin, improving FADE by 0.241 and BRISQUE by 6.95.
On Fattal’s dataset, our HazeFlow outperforms the next leading method by a margin of 0.074 in FADE score, while also obtaining best results in BRISQUE and PAQ2PIQ.

Next, for evaluation on the paired real-world Dense-HAZE and NH-HAZE datasets, we compare our method against SOTA approaches, including the prior-based method DCP~\cite{he2010single}, methods trained on the RESIDE~\cite{li2018benchmarking} dataset (AOD-Net~\cite{li2017aod}, MSBDN~\cite{dong2020multi}, GCANet~\cite{das2022gca}, FFANet~\cite{qin2020ffa}, DehazeFormer~\cite{song2023vision}, Dehamer~\cite{guo2022dehamer}), and diffusion-based methods (Wang \textit{et al.}~\cite{wang2024frequency}, DehazeDDPM~\cite{yu2023high}).  
For DehazeDDPM, we use the official pretrained weights for evaluation, while for Wang \textit{et al.}, we fine-tune the model on Dense-HAZE and NH-HAZE using the same number of iterations as HazeFlow to ensure a fair comparison.
In Tab.~\ref{tab:quantitativepaired}, HazeFlow outperforms all competing methods in Dense-HAZE, achieving the best scores across all metrics. 
Specifically, HazeFlow outperforms the second-best method by a substantial margin, with a 0.0556 improvement in SSIM and 0.1530 in LPIPS.
On the NH-HAZE dataset, HazeFlow achieves the best perceptual quality, ranking first in LPIPS and SSIM while securing the second-highest PSNR.
Notably, our method produces perceptually superior outcomes relative to SOTA diffusion-based methods (DehazeDDPM), while delivering competitive objective performance against the regression-based SOTA approach Dehamer. 
These findings demonstrate the effectiveness of our approach in addressing complex real-world haze, efficiently achieving a balance between restoration quality and structural integrity.



\begin{table}[t]
    \centering\resizebox{\columnwidth}{!}{
    \begin{tabular}{l|ccc|ccc}
    \toprule
    \multirow{2}{*}{Method} & \multicolumn{3}{c}{Dense-HAZE~\cite{ancuti2019dense}} & \multicolumn{3}{c}{NH-HAZE~\cite{ancuti2020nh}} \\
    \cmidrule(r){2-4} \cmidrule(r){5-7}
    & PSNR$\uparrow$ & SSIM$\uparrow$ & LPIPS$\downarrow$ & PSNR$\uparrow$ & SSIM$\uparrow$ & LPIPS$\downarrow$ \\
    \midrule
    DCP~\cite{he2010single}             & 11.01  &0.4165 &0.8405 &12.72 &0.4419 &0.5168 \\
    AOD-Net~\cite{li2017aod}            & 12.88  &0.5045 &0.7612 &15.31 &0.4584 &0.5121 \\
    MSBDN~\cite{dong2020multi}          & 13.44  &0.4301 &0.9501 &17.34 &0.5566 &0.5026 \\
    GCANet~\cite{das2022gca}            & 12.46  &0.4774 &0.7313 &16.64 &0.5583 &0.4356 \\
    FFANet~\cite{qin2020ffa}           & 15.18  &0.5757 &0.6072 &18.48 &0.6186 &0.3694 \\
    DehazeFormer~\cite{song2023vision}  & 14.95  &0.4979 &0.7770 &18.15 &0.6070 &0.4192 \\
    Dehamer~\cite{guo2022dehamer}       & \underline{16.62} & \underline{0.5602} & 0.6543 & \textbf{20.66} & \underline{0.6844} &0.3122 \\
    \midrule
    Wang \textit{et al.}~\cite{wang2024frequency}                & 14.83  & 0.4064 & 0.7508 & 16.11 & 0.5591 & 0.4349\\
    DehazeDDPM~\cite{yu2023high}        & 15.45  & 0.4559 & \underline{0.5808} & 19.44 & 0.6278 & \underline{0.2984} \\
    HazeFlow                            & \textbf{17.14} & \textbf{0.6158} & \textbf{0.4278} & \underline{20.06} & \textbf{0.6847} & \textbf{0.2606} \\
    \bottomrule
    \end{tabular}}
    \vspace{-1.5ex}
\caption{Quantitative results on paired real-world datasets: Dense-HAZE~\cite{ancuti2019dense} and NH-HAZE~\cite{ancuti2020nh}. Best results are \textbf{bolded} and second best results are \underline{underlined}. The last three rows correspond to generative model-based approaches.}
\vspace{-1.5em}
\label{tab:quantitativepaired}
\end{table}

\vspace{1ex}
\noindent\textbf{Qualitative Results.}
We provide qualitative comparisons on both unpaired~(Fig.~\ref{fig:unpaired}) and paired datasets (Fig.~\ref{fig:paired}). 
As shown in Fig.~\ref{fig:unpaired}, our method excels in restoring background regions with significant depth. 
The first row shows no residual haze, while the second row highlights effective haze removal from distant objects, as marked.
Similarly, Fig.~\ref{fig:paired} shows that HazeFlow achieves more precise haze removal and color prediction results, closely matching ground truth images.
Notably, our model accurately captures the spatial variant in haze and dynamically adjusts its dehazing strength, leading to sharper, artifact-free results across varying haze densities.
More results can be found in the Supplementary Material Sec.~\ref{sup:additional_results}.

\begin{table*}[t]
    \centering
    \resizebox{0.99\textwidth}{!}{
    \begin{tabular}{l|cccc|cccc|cccc}
        \hline
        \multirow{2}{*}{Method}& \multicolumn{4}{c|}{DehazeFormer~\cite{song2023vision}} & \multicolumn{4}{c|}{NAFNet~\cite{chen2022nafnet}} & \multicolumn{4}{c}{FocalNet~\cite{cui2023focal}} \\
        \cmidrule(r){2-5} \cmidrule(r){6-9} \cmidrule(r){10-13}
        & FADE$\downarrow$ & BRISQUE$\downarrow$ & NIMA$\uparrow$ & PAQ2PIQ$\uparrow$
        & FADE$\downarrow$ & BRISQUE$\downarrow$ & NIMA$\uparrow$ & PAQ2PIQ$\uparrow$ 
        & FADE$\downarrow$ & BRISQUE$\downarrow$ & NIMA$\uparrow$ & PAQ2PIQ$\uparrow$ \\
        \hline
        DAD$*$~\cite{shao2020dad} & 6.16 & 45.45 &  4.93 & 64.74 & 2.82 & 36.88 & 4.80 & 66.52 & 2.69 & 33.68 &4.82& 66.77 \\
        D4$*$~\cite{yang2022self} & 1.78 & 29.60 &4.85& 66.71 & 1.35 & 28.64 & 4.81 & 67.07 & 1.16 & 28.78 & 4.86 & 67.21 \\
        RIDCP~\cite{wu2023ridcp} & 1.44 & 28.43 &5.17& 67.96 & 1.17 & \textbf{21.01} & \textbf{5.00} & 67.48 & 1.36 & \textbf{28.07} & 4.99 & 65.47 \\
        \hline
        MCBM~(Ours) & \textbf{1.05} & \textbf{23.22} &\textbf{5.25}& \textbf{68.40} & \textbf{1.03} & 23.68 &4.94& \textbf{67.63} & \textbf{0.80} & 29.90 &\textbf{5.06}& \textbf{68.44} \\
        \hline
    \end{tabular}
    }
    \vspace{-0.75em}
    \caption{Comparison of different haze augmentation methods applied to baseline models. Methods marked with $*$ (DAD and D4) require training, while RIDCP and MCBM (Ours) are training-free approaches. Best results are highlighted in \textbf{bold}.}
    \vspace{-1.5em}
    \label{tab:MCBM_comparison}
\end{table*}

\subsection{Comparison with Diffusion-based Methods}
To demonstrate the superior efficiency and effectiveness of our generative approach, we compare it with recent diffusion-based methods~\cite{wang2024frequency,luo2023refusion}, including all-in-one models such as DiffUIR~\cite{zheng2024diffuir} and DACLIP~\cite{luo2023daclip}, which are trained on large-scale datasets incorporating various types of degradation.
Additionally, we compare against Refusion~\cite{luo2023refusion}, which is trained on non-homogeneous real pairs~\cite{ancuti2023ntire}, as well as the dehazing model by Wang \textit{et al.}~\cite{wang2024frequency}, which is trained on the RESIDE dataset. 
As shown in Tab.~\ref{tab:compare_diffusion}, HazeFlow achieves significantly better performance with substantially fewer steps, leading to much faster runtime.
This highlights that solving the ODE of ASM with a flow-based approach enhances both speed and restoration quality.

\subsection{MCBM Performance Across Different Models}
To illustrate the broad applicability of MCBM, we assess its performance with various baselines, such as NAFNet~\cite{chen2022nafnet}, DehazeFormer~\cite{song2023vision}, and FocalNet~\cite{cui2023focal}, as shown in Tab.~\ref{tab:MCBM_comparison}.
The results show that MCBM consistently achieves the best performance in FADE and PAQ2PIQ across all baselines, confirming its effectiveness in training various models. BRISQUE show best performance with DehazeFormer, and NIMA show better performance with FocalNet.
Notably, our MCBM is the only approach that is both training-free and capable of generating non-homogeneous haze, unlike methods such as DAD~\cite{shao2020dad}, D4~\cite{yang2022self}, and RIDCP~\cite{wu2023ridcp}.
This indicates that MCBM haze generation process closely aligns with real-world haze characteristics, making it adaptable to various models.

\subsection{Ablations}
To evaluate HazeFlow, we conduct experiments on the three-stage learning framework, MCBM and T-Refinement. 
All ablation studies are performed on the RTTS dataset~\cite{li2018benchmarking}.

\vspace{1ex}
\noindent\textbf{Effectiveness of Three-Stage Learning.}
Tab.~\ref{tab:abl_2} presents an ablation study on Reflow and Distillation, highlighting their effectiveness and showing significant performance improvements.
Specifically, Reflow achieves substantial gains in BRISQUE and PAQ2PIQ, demonstrating its ability to adapt to real-world datasets and accurately learn complex haze distributions.
Finally, Distillation further enhances performance for real-world degradation in terms of BRISQUE, PAQ2PIQ, and NIMA. 
Supplementary Material Sec.~\ref{sec:supp_reflow and distill} provide additional visual results of comparison between Reflow and Distillation.

\vspace{1ex}
\noindent\textbf{Effectiveness of MCBM and T-Refinemet.}
Tab.~\ref{tab:abl_1} shows that HazeFlow trained with our MCBM outperforms models trained with previous haze synthesis approach~\cite{wu2023ridcp}, improving the FADE score by approximately 0.133.
This demonstrate that MCBM significantly enhances real-world haze removal.
Additionally, T-Refinement corrects inaccuracies in the transmission map predicted by DCP, enabling HazeFlow to estimate transmission more accurately.
This refinement improves the FADE score by 0.234 and further enhances haze removal.
Adding T-Refinement also enhances the recovery of color and brightness restoration during inference.
Visual results for T-Refinement are provided in the Supplementary Material Sec.~\ref{sec:supp_transmission refinement}.

\begin{table}[t]
    \centering
    \resizebox{\columnwidth}{!}{
    \begin{tabular}{l|cccc|c|c}
        \hline
         Method& FADE$\downarrow$ & BRISQUE$\downarrow$ & NIMA$\uparrow$ & PAQ2PIQ$\uparrow$ &  Run time (s) & \# of steps \\
        \hline
        DACLIP~\cite{luo2023daclip}  & 1.922 & 34.78 &4.98  & 66.87 & 38.09 & 100 \\
        Wang \textit{et al.}~\cite{wang2024frequency}& 0.929 & 27.66 & 5.00 & 67.51 & 2.50 & 20 \\
        DiffUIR~\cite{zheng2024diffuir}  & 2.019 & 34.88 &4.54  & 66.81 & 0.30 & 3 \\
        Refusion~\cite{luo2023refusion} & 2.047 & 42.59 & 4.49 & 64.67 & 2.76 & 100 \\
         \hline
        HazeFlow & \textbf{0.583} & \textbf{5.01} & \textbf{5.30} & \textbf{72.97} & \textbf{0.21} & {\textbf{1}} \\
        \hline
    \end{tabular}}
    \vspace{-2.0ex}
    \caption{Comparison with other diffusion-based models.} 
    \label{tab:compare_diffusion}
    \vspace{-2.0ex}
\end{table}

\begin{table}[t]
    \centering\resizebox{\columnwidth}{!}{
    \begin{tabular}{l|c|cccc}
      \toprule
      \multicolumn{2}{c|}{Experiment} & FADE$\downarrow$ & BRISQUE$\downarrow$ & NIMA$\uparrow$ & PAQ2PIQ$\uparrow$ \\
      \midrule
      Pretrained    & $h_\theta$ &0.561&18.96&5.26 &	69.58\\ 
      \midrule
      + Reflow      & $h_\phi$   &0.598&8.14 &5.26&	72.49\\ 
      \midrule
      + Distillation& $h_{\phi'}$&{0.583}&{5.01} &{5.30}& {72.97}\\ 
    \bottomrule
    \end{tabular}}
    \vspace{-2.0ex}
    \caption{Ablation study on 3-stage framework.}
    \vspace{-2.0ex}
    \label{tab:abl_2}
\end{table}

\begin{table}[t]
    \Large \centering\resizebox{\columnwidth}{!}{
    \begin{tabular}{cc|cccc}
      \toprule
       MCBM & T-Refinement & FADE$\downarrow$ & BRISQUE$\downarrow$ & NIMA$\uparrow$ & PAQ2PIQ$\uparrow$ \\
      \midrule
           -     &     -      & 0.950&14.63&5.31 &70.78 \\ 
      \checkmark &     -      & 0.817&12.67&5.35 &71.38  \\
      \checkmark & \checkmark & {0.583}& {5.01}&{5.30} &{72.97} \\ 
    \bottomrule
    \end{tabular}}
    \vspace{-1.5ex}
    \caption{Ablation study for MCBM and Transmission refinement.}
    \vspace{-1.0em}
    \label{tab:abl_1}
\end{table}

%% file: sec/6_conclusion.tex
\vspace{-0.5em}
\section{Conclusion}
\vspace{-0.5em}
In this work, we present HazeFlow, a novel framework that unifies physics-guided modeling and neural ODEs to redefine single image dehazing.
By reformulating the ASM as an ODE, we transform dehazing into a trajectory optimization problem, where a neural network learns a straight ODE path from atmospheric light to the clean image, guided by the transmission map. 
This design not only enhances dehazing accuracy but also enables fast inference, achieving high-quality results even with a single step while drastically reducing computational costs compared to iterative methods.
Furthermore, by modeling non-homogeneous haze synthesis with MCBM, we enhance our framework's capacity to handle complex and diverse real-world hazy scenes. 
This comprehensive approach aims to bridge the gap between synthetic and real-world data, improving generalization and robustness while accelerating inference and advancing dehazing performance.

%% file: sec/acknowlege.tex
\section*{Acknowledgement}
\vspace{-0.5em}

This work was supported by National Research Foundation of Korea (NRF) grant funded by the Korea government (MSIT) (RS-2023-00222776), and Institute of Information communications Technology Planning Evaluation (IITP) grant funded by the Korea government (MSIT) (No.2022- 0-00156, Fundamental research on continual meta-learning for quality enhancement of casual videos and their 3D metaverse transformation) and IITP grant funded by the Ministry of Science and ICT (No.RS-2020-II201373, Artificial Intelligence Graduate School Program [Hanyang University]).

\vspace{-0.5em}

%% file: sec/X_suppl.tex
\clearpage
\setcounter{page}{1}
\setcounter{section}{0}
\maketitlesupplementary

\renewcommand{\thesection}{S\arabic{section}}
\renewcommand{\thesubsection}{S\arabic{section}.\arabic{subsection}}  




\section{Reflow and Distillation}

\begin{figure}[h]
    \centering
    \includegraphics[width=\linewidth]{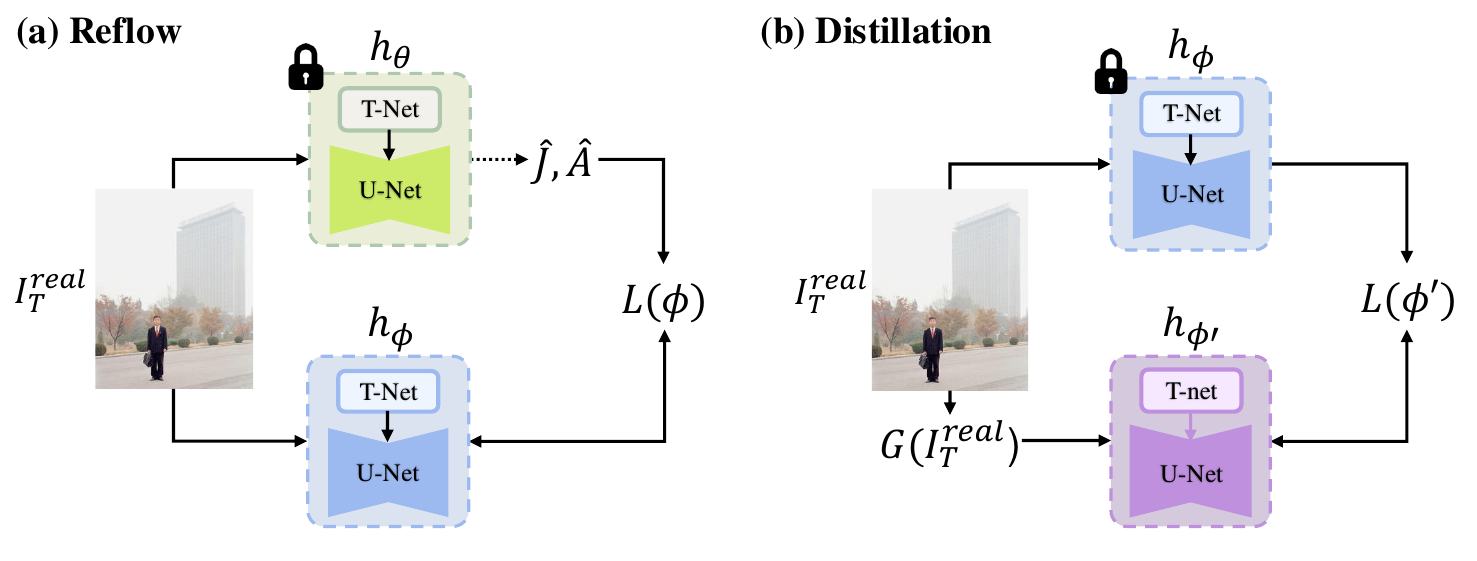}
    \caption{Illustration of (a)~Reflow and (b)~Distillation in our 3-stage learning framework.}
    \vspace{-1.em}
    \label{fig:reflow_distill}
\end{figure}

For a more detailed understanding, we provide illustrations depicting the Reflow and Distillation processes Fig~\ref{fig:reflow_distill}.

\subsection{Implementation Details}
In Reflow and Distillation phase, We train HazeFlow for 10K iterations on the URHI~\cite{li2018benchmarking}, using Adam~\cite{kingma2014adam} optimizer.
We use learning rate of $1 \times 10^{-6}$ for Reflow and $5 \times 10^{-5}$ for Distillation.
Other details are consistent with those described in the pretraining phase.

\subsection{Additional Visual Results}
\label{sec:supp_reflow and distill}
\begin{figure*}
    \centering
    \includegraphics[width=\linewidth]{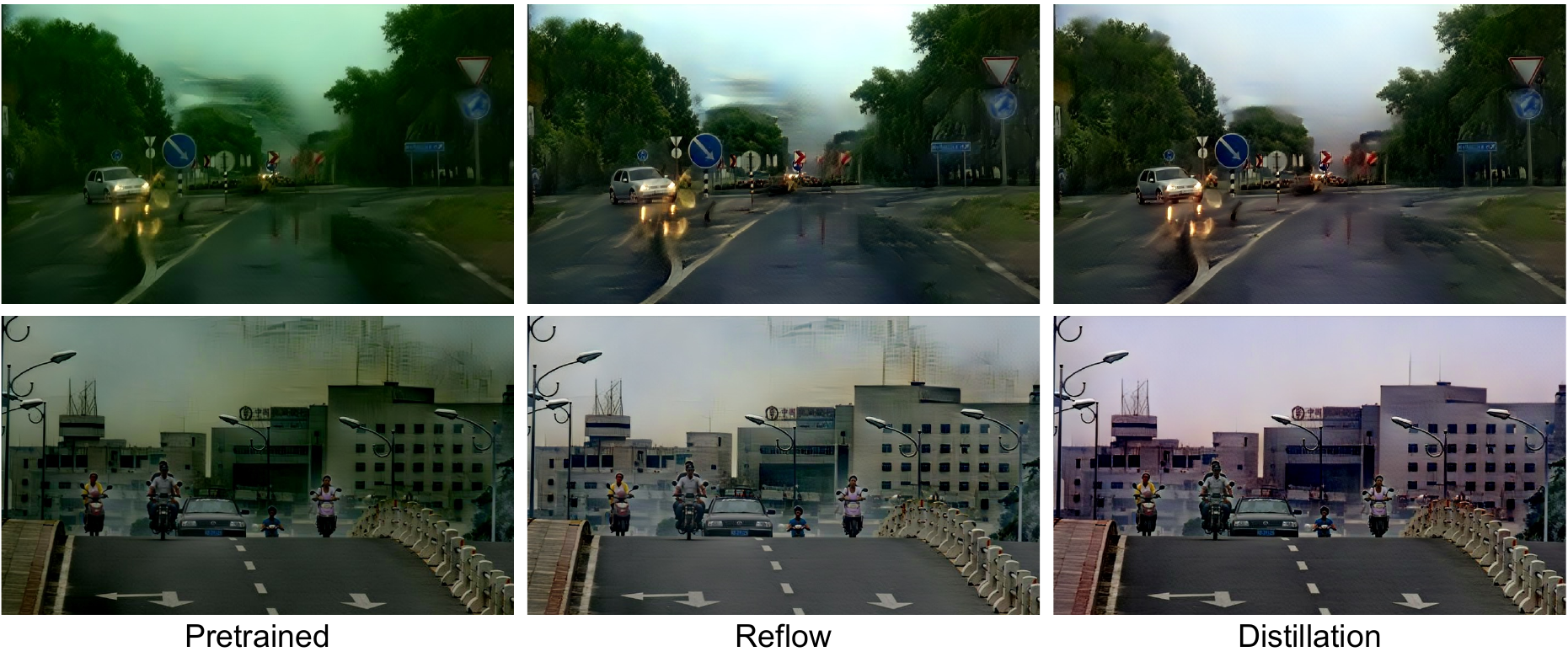}
    \caption{Visual comparison between the three phases of our method on the RTTS dataset~\cite{li2018benchmarking}.}
    \label{fig:supp_pre-re-distil}
\end{figure*}

For a phase-by-phase comparison, we present visual results on the RTTS dataset~\cite{li2018benchmarking} in Fig.~\ref{fig:supp_pre-re-distil}.
During the pretraining phase, the network focuses primarily on learning to remove haze effectively.
In the Reflow phase, the network is trained with pseudo atmospheric light $\hat{A}$, enabling it to restore appropriate lighting in hazy regions effectively.
In the Distillation phase, as shown in Fig.~\ref{fig:supp_pre-re-distil}, this stage removes artifacts and enhances the natural appearance of the results.

\section{Transmission Refinement}
\label{sec:supp_transmission refinement}
We utilize the refined transmission map $T_{refined}$ across multiple stages, including perceptual loss calculation and sampling during inference, Reflow and Distillation.
Specifically, the perceptual loss described in Eq.~\ref{eq:lpips}, utilizing the refined transmission map, can be written as follows:
\begin{equation}
    \label{eq:lpips_refined}
    L_{perc}(\theta) = \mathbb{E}[\mathbb{D}(J,\ I_T + (1-T_{refined})h_\theta(I_T, T_{DCP}))].
\end{equation} 
Similarly, the inference process, as described in Eq.~\ref{eq:ode_solver}, can be expressed as:
\begin{equation}
    I_{T+ {(1-T)}\frac{1}{N}} = I_T + (1-T_{refined}) \frac{1}{N} h_\theta(I_T,T_{DCP}).
\end{equation}
The obtained transmission maps $T_{refined}$ are also utilized for sampling pseudo images during the Reflow phase.
The sampling process for the pseudo clean image $\hat{J}$ and pseudo atmospheric light $\hat{A}$ can be expressed as:
\begin{align}
    &\hat{J} = I^{real}_T + (1-T_{refined})\cdot h_\theta(I^{real}_T, T_{DCP}),  \\
    &\hat{A} = I^{real}_T - T_{refined}\cdot h_\theta(I^{real}_T, T_{DCP}).
\end{align}
Also, in Distillation phase, we use refined transmission map $T_{refined}$ instead of transmission map obtained by DCP.

\begin{align}
    L_{Distill}(\phi') =  
    &\mathbb{E} \big[ \mathbb{D}(I_T + (1 - T_{refined}) h_\phi(I_T, T_{DCP}), \\
    & \mathcal{G}(I_T) + (1 - T_{refined}) h_{\phi'}(\mathcal{G}(I_T), T_{DCP})) \big].
\end{align}

The refined transmission map $T_{refined}$ provides a more accurate estimation of haze density compared to $T_{DCP}$. 
As a result, training with $T_{refined}$ enables the model to remove haze more effectively and accurately.

\begin{figure*}
    \centering
    \includegraphics[width=\linewidth]{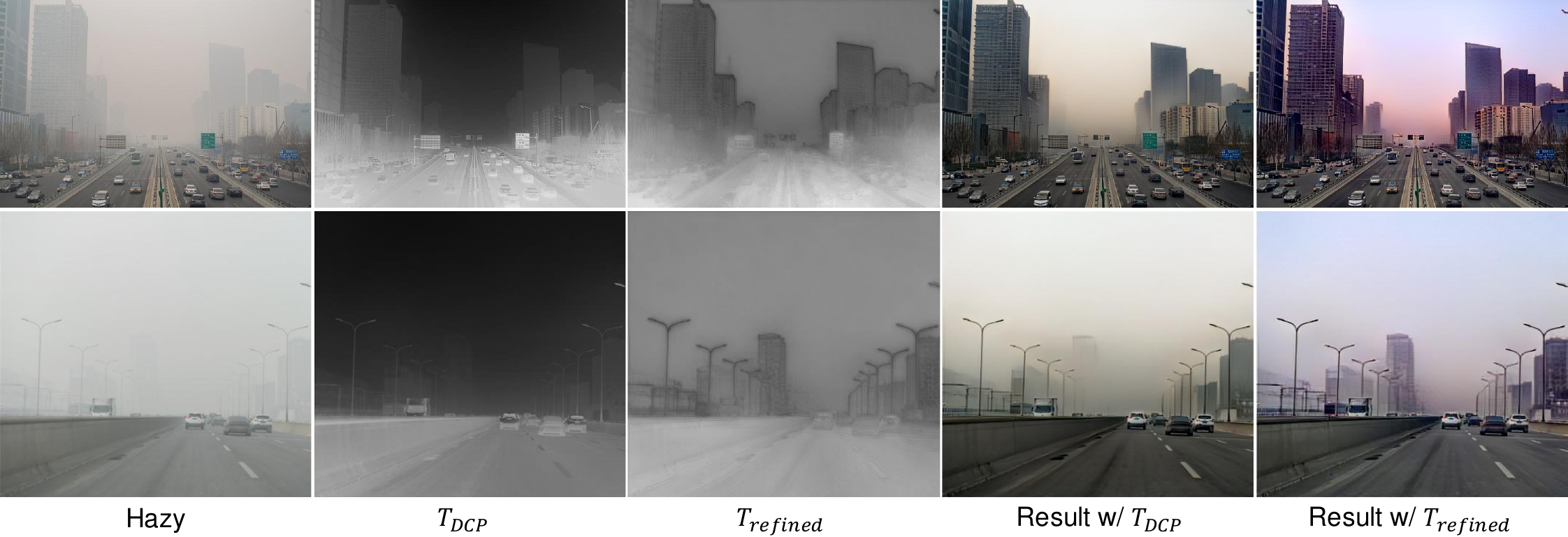}
    \caption{Visual comparison between results with $T_{DCP}$ and with $T_{refined}$ on the RTTS dataset~\cite{li2018benchmarking}.}
    \label{fig:supp_refine}
\end{figure*}

As shown in Fig.~\ref{fig:supp_refine}, our refined transmission map captures haze more effectively.
DCP struggles to capture haze in regions with high depth, leaving residual haze in such areas, and it also fails to effectively remove haze in sky regions due to its limitations. 
In contrast, our method accurately captures haze density, enabling effective haze removal not only in high-depth regions but also in sky areas.

\section{MCBM}
\label{supp:mcbm}
\subsection{Implementation Details}
\begin{figure}[h]
    \centering
    \includegraphics[width=\linewidth]{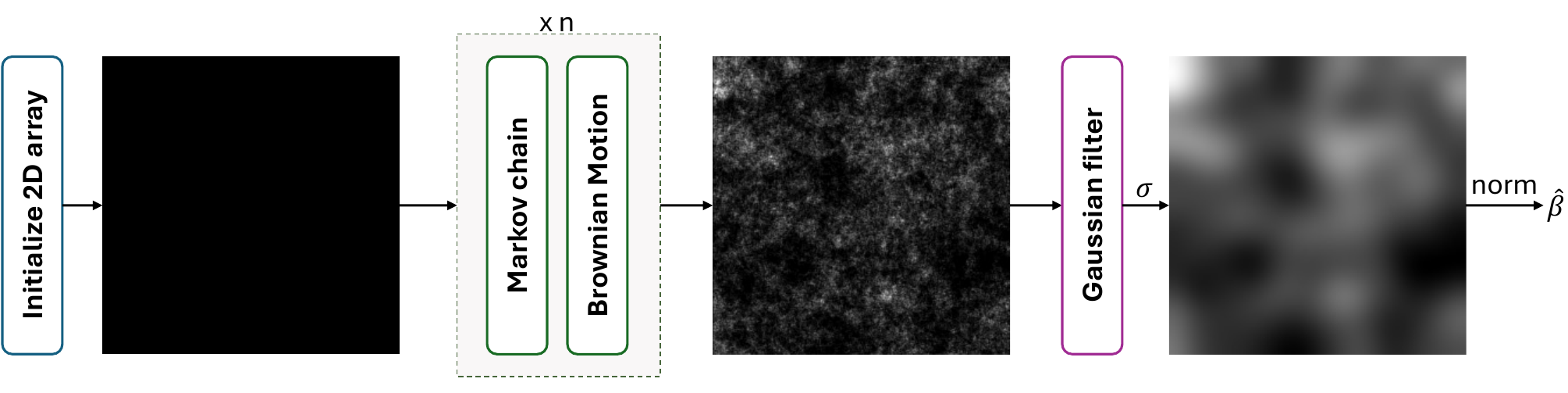}
    \caption{Overall pipeline of MCBM.}
    \label{fig:supp_mcbm_pipeline}
\end{figure}

For a more detailed explanation of MCBM haze generation process, the overall pipeline is illustrated in Fig.~\ref{fig:supp_mcbm_pipeline}. 
To obtain diverse shapes of non-homogeneous haze density, we treat the number of iterations $n$ and the strength of the Gaussian filter $\sigma$ as hyperparameters. 
Specifically, the iteration $n$ represents the number of times the Markov chain and Brownian motion are performed, and a higher $n$ leads to reduced non-homogeneity.
We randomly select $n$ by multiplying the total number of pixels by factors of $[4, 5, 6]$.
Additionally, to create realistic haze density, it is necessary to smooth to the 2D array generated by the Markov chain and Brownian motion. 
We apply smoothing using a Gaussian filter, where the strength $\sigma$ is randomly selected from $[15, 25, 35]$.
Finally, the MCBM haze density is generated through normalization.

\subsection{Additional Visual Results}
\begin{figure*}
    \centering
    \includegraphics[width=0.8\linewidth]{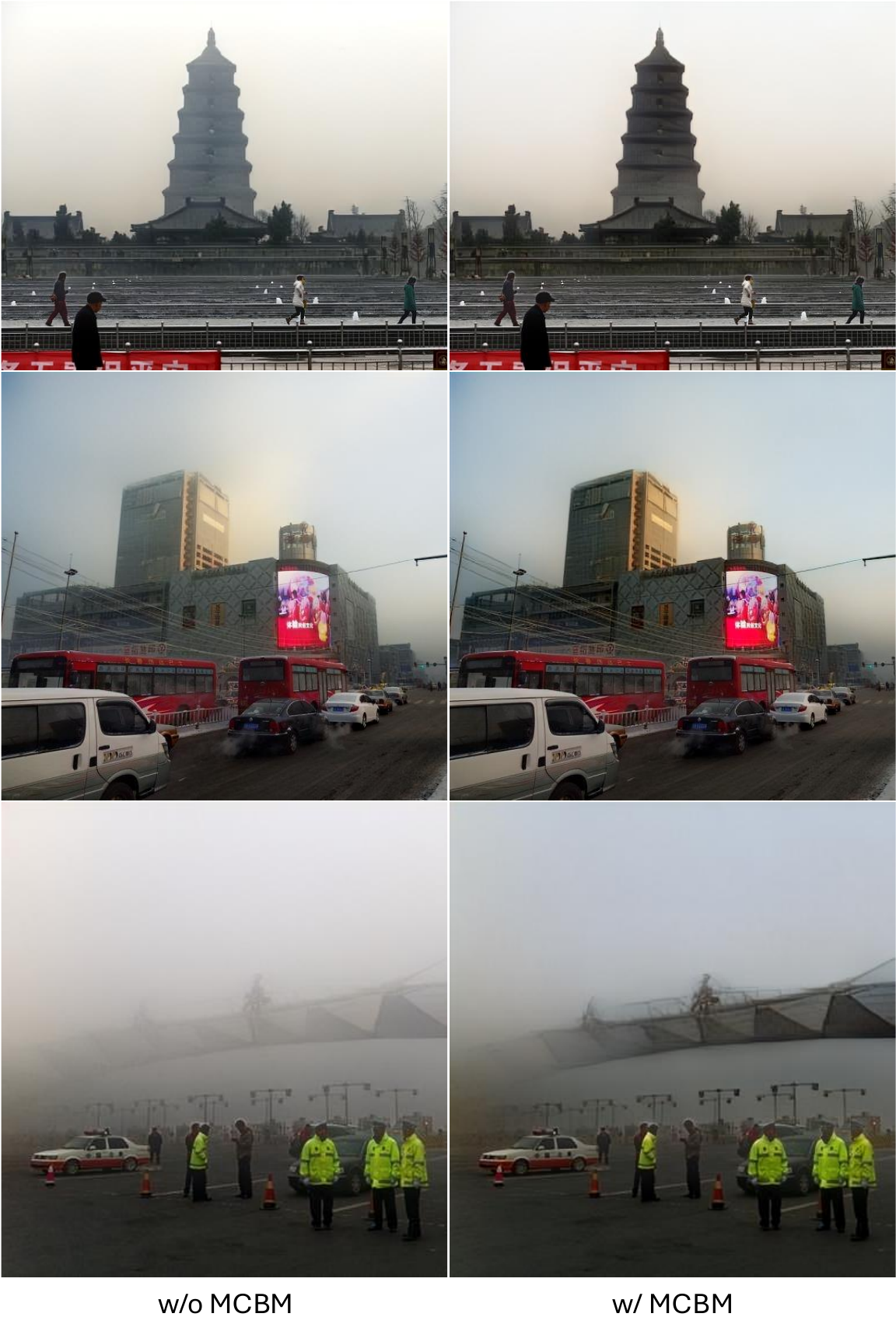}
    \caption{Visual comparison of the results from networks trained with and without MCBM haze on the RTTS dataset~\cite{li2018benchmarking}.}
    \label{fig:supp_mcbm}
\end{figure*}

To demonstrate the effectiveness of our MCBM haze synthesis, we provide additional visual results.
For a fair comparison, we evaluate both networks without applying our transmission refinement process on the RTTS~\cite{li2018benchmarking}.
As shown in Fig.~\ref{fig:supp_mcbm}, the network trained with our MCBM haze synthesis produces significantly clearer results.
By learning to capture non-homogeneous haze, the network effectively removes haze more comprehensively.

\subsection{t-SNE Visualization of MCBM Haze}

\begin{figure}[h]
    \centering
    \includegraphics[width=0.7\linewidth]{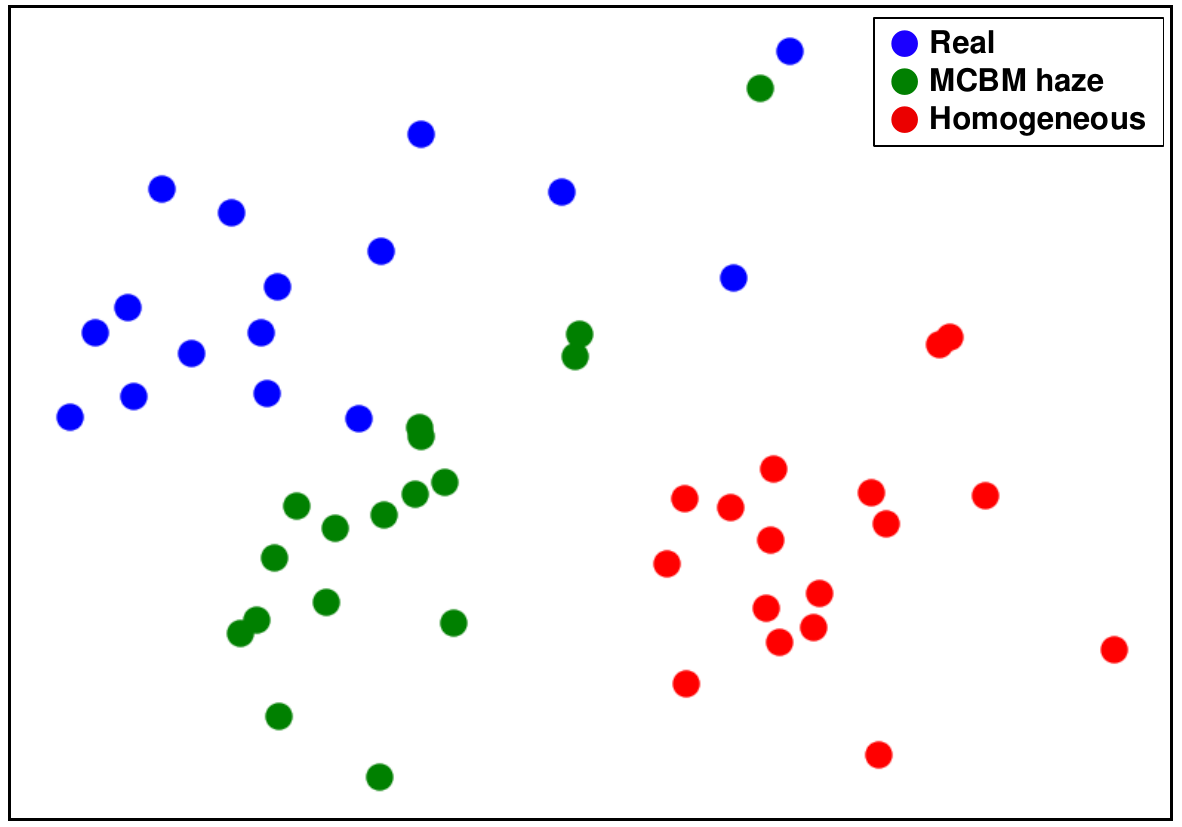} 
    \caption{Visualization of feature distributions of pretrained network~\cite{simonyan2014vgg} using t-SNE, comparing real hazy images (\textcolor{blue}{blue}) obtained from NH-HAZE~\cite{ancuti2020nh}, synthesized non-homogeneous hazy images (\textcolor{darkgreen}{green}) generated by our MCBM haze density, and synthesized homogeneous hazy images (\textcolor{red}{red}).}

    \label{fig:tsne}
\end{figure}

To verify whether MCBM can closely approximate real-world hazy images, we provide a visualization using t-SNE. 
As shown in Fig.~\ref{fig:tsne}, non-homogeneous hazy images generated by MCBM haze density is much closer to real-world hazy image in the feature space of VGG networks~\cite{simonyan2014vgg}, compared to homogeneous hazy image.
This approach enables a closer approximation to real haze conditions and the resulting synthetic images facilitate the effective learning of our dehazing networks for real-world scenarios.

\section{User Study}
\begin{table}[h]
    \centering\resizebox{1.0\columnwidth}{!}{
    \begin{tabular}{l|cccccc}
      \toprule
      Method &DAD~\cite{shao2020dad} & PSD~\cite{chen2021psd} & D4~\cite{yang2022self}& RIDCP~\cite{wu2023ridcp} & CORUN~\cite{fang2024real} & HazeFlow \\
      \midrule
      Score$\uparrow$  &0.0273& 0.0076& 0.0440& 0.1214& 0.1730& \textbf{0.6267}\\
    \bottomrule
    \end{tabular}}
    \vspace{-1.em}
    \caption{User study result on RTTS and Fattal's datasets.} 
    \vspace{-0.5em}
    \label{tab:user_study}
\end{table}

To provide a more thorough comparison, we conduct a user study. 
In this study, we randomly select 4 images from Fattal’s dataset and 26 images from the RTTS dataset. 
Participants are asked to evaluate the results based on three criteria: (1) the completeness of haze removal, (2) the absence of artifacts, and (3) the quality of color restoration. 
They then chose best performing image from the results of our method and those of the state-of-the-art models.
The study involved 5 image processing experts and 17 non-expert participants.
As shown in Tab.~\ref{tab:user_study}, our method received the majority of votes, with a significant margin over the second-ranked model. 
This result demonstrates that our HazeFlow effectively removes haze, even according to human perception.

\section{Discussion on Estimation Steps}
\label{sup:abl_steps}
\begin{figure}[h]
    \centering
    \includegraphics[width=\linewidth]{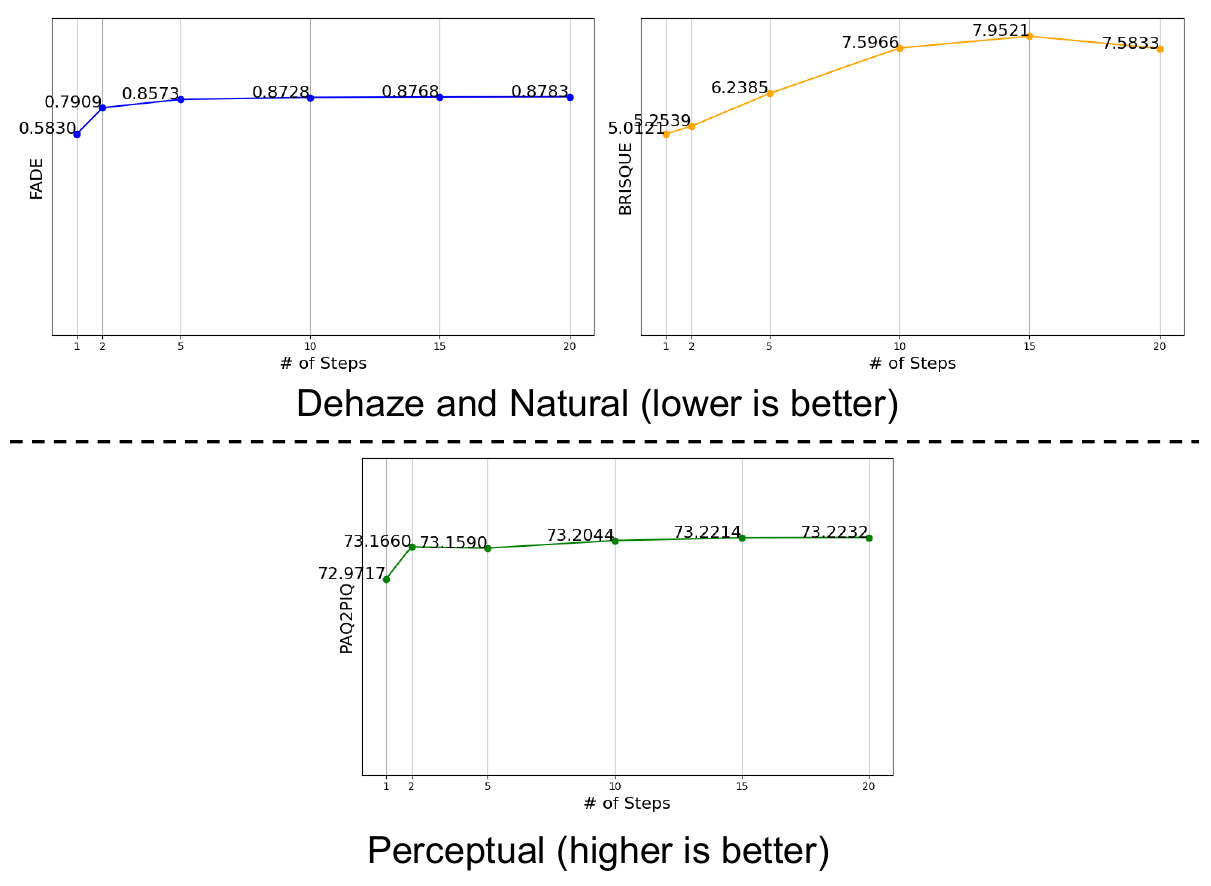}
    \caption{Variation in metrics based on the number of estimation steps.}
    \label{fig:supp_metrics}
\end{figure}
\begin{figure}[h]
    \centering
    \includegraphics[width=\linewidth]{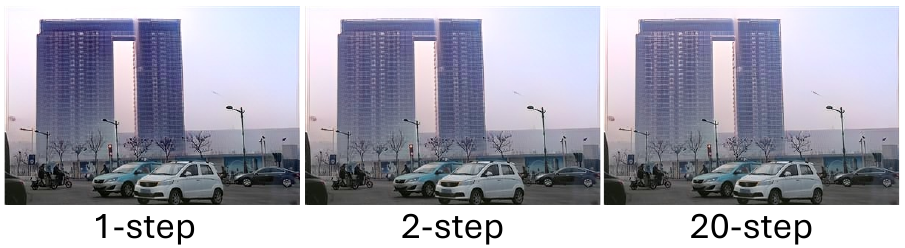}
    \caption{Visualization of results with varying numbers of estimation steps.}
    \label{fig:vis_steps}
\end{figure}

HazeFlow can predict results using multiple estimation steps, similar to other RF-based models~\cite{liu2022flow, wang2024semflow, zhu2024flowie}.
This section explains the impact of the number of estimation steps on the results.
For thorough evaluation, we use four metrics: FADE~\cite{choi2015referenceless} and BRISQUE~\cite{mittal2012brisque} assess how well the predicted image has been dehazed and how close it is to the natural image, while PAQ2PIQ~\cite{ying2020paq2piq} evaluates the perceptual quality of the predicted images. 
As shown in Fig.~\ref{fig:supp_metrics}, as the number of steps increases, FADE and BRISQUE scores degrade, while PAQ2PIQ scores improve. 
This indicates that there is a trade-off between the degree of dehazing and the perceptual quality in the predicted image. 
Although increasing the number of estimation steps may enhance perceptual quality, this study focuses on achieving effective dehazing while minimizing computational cost.
Consequently, one-step estimation is chosen as the final approach.
Predicted images for different estimation steps can be seen in Fig.~\ref{fig:vis_steps}. 

\section{Comparison with Rectified Flow}
\begin{figure*}
    \centering
    \includegraphics[width=0.8\linewidth]{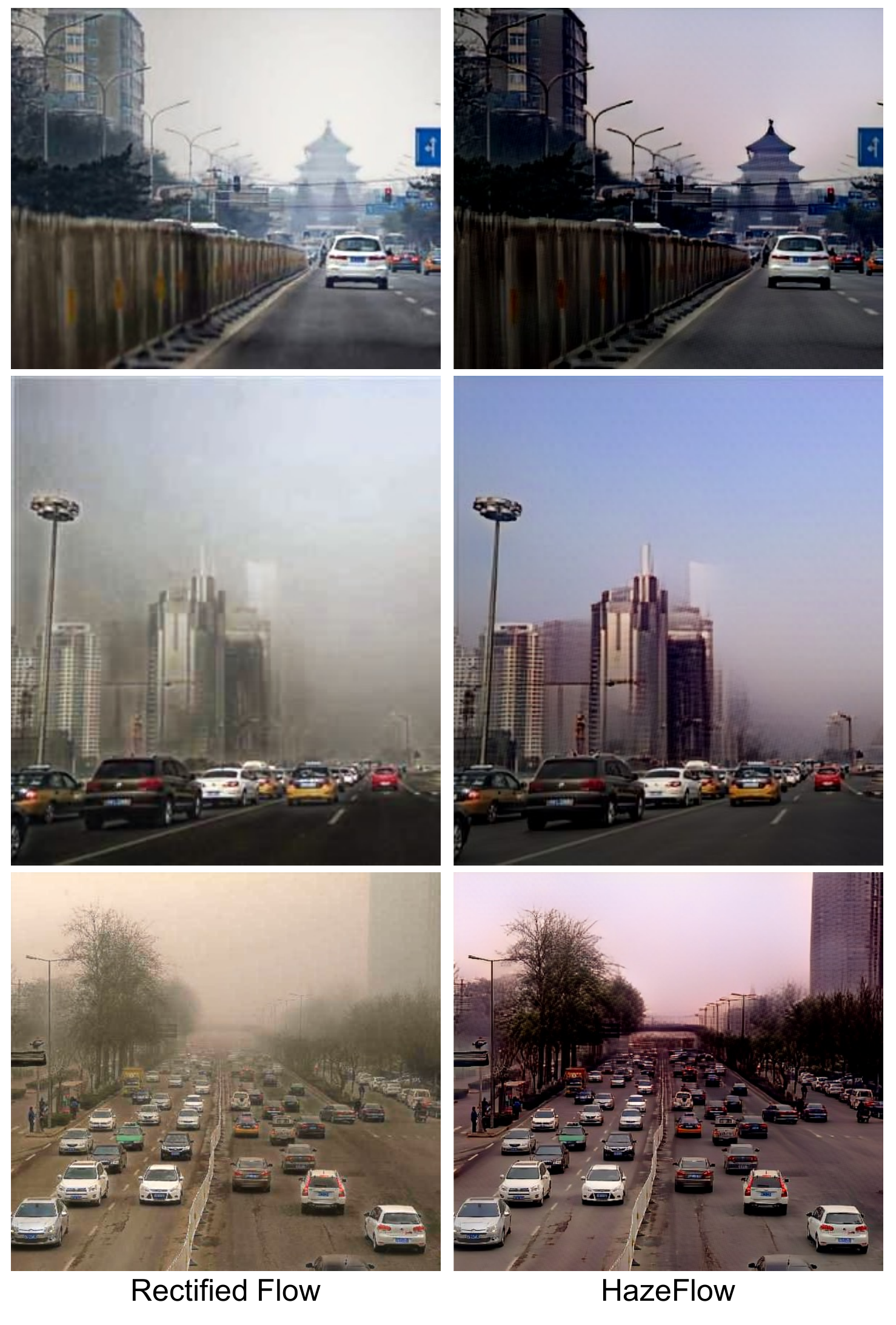}
    \caption{Visual comparison between Rectified Flow~\cite{liu2022flow} and HazeFlow (Ours) on RTTS dataset~\cite{li2018benchmarking}.}
    \label{fig:supp_RFvsHF}
\end{figure*}

In this section, we demonstrate that the ASM-based ODE outperforms the naive ODE derived from linear interpolation by comparing it with the baseline, RF~\cite{liu2022flow}.
To train RF, we assume the hazy distribution as $X_0$ and the clean distribution as $X_1$ in Sec.~\ref{sec:4.1}.
For a fair comparison, Reflow and Distillation are conducted as Eq.~(\ref{eq:our_reflow}) and Eq.~(\ref{eq:our_distill}) using one-step estimation. 
We provide a quantitative comparison in Tab.~\ref{tab:supp_RFvsHF}, which shows that HazeFlow outperforms RF in both dehazing capability and perceptual quality.
Additionally, a visual comparison is provided in Fig.~\ref{fig:supp_RFvsHF}.
\begin{table}[h]
    \centering\resizebox{1.0\columnwidth}{!}{
    \begin{tabular}{l|cccc}
      \toprule
      Method & FADE$\downarrow$ & BRISQUE$\downarrow$ & PAQ2PIQ$\uparrow$ & MUSIQ$\uparrow$ \\
      \midrule
      Rectified Flow &1.059&32.21&69.81&	54.16\\ 
      \midrule
      HazeFlow &\textbf{0.583} &\textbf{5.01}&	\textbf{72.97}&	\textbf{63.94} \\
    \bottomrule
    \end{tabular}}
    \caption{Quantitative comparison between RF~\cite{liu2022flow} and our HazeFlow.}
    \label{tab:supp_RFvsHF}
\end{table}

\section{Other Transmission Estimation Methods}
\begin{table}[b]
    \centering
    \resizebox{\columnwidth}{!}{
    \begin{tabular}{l|cccc}
        \hline
        Method & FADE$\downarrow$ & BRISQUE$\downarrow$ & NIMA$\uparrow$ & PAQ2PIQ$\uparrow$ \\
        \hline
        DCP~\cite{he2010single} & 0.583 & 5.01 & 5.30 & 72.97 \\
        DCPDN~\cite{zhang2018dcpdn} & 0.713 & 4.94 &  5.19& 71.95 \\
        Non-local~\cite{berman2016non_local} & 0.734 & 1.38 &5.16  & 73.11 \\
        \hline
    \end{tabular}
    }
    \caption{Ablation on different transmission estimation methods.}
    \label{tab:transmission_est}
\end{table}

Although DCP~\cite{he2010single} is used to approximate transmission map in our method, various alternative transmission estimation methods can also be employed.
Tab.~\ref{tab:transmission_est} compares different transmission estimation methods including DCP~\cite{he2010single}, the prior-based method Non-Local~\cite{berman2016non_local} and the model-based method DCPDN~\cite{zhang2018dcpdn}.
We selected DCP due to its high FADE score to enhance dehazing quality, but other methods can be chosen depending on the objective. 
Note that selecting the initial transmission map offers flexibility and the potential for further performance improvements through alternative approaches.

\section{Additional Quantitative Results in paired datasets.}
\label{sup:paired}
We also conduct a comparison without additional training on paired datasets.
As shown in Tab.~\ref{tab:supp_paired}, HazeFlow achieves the best performance in both PSNR and SSIM across all datasets except Dense-HAZE.
HazeFlow achieves improvements of 0.43 dB in PSNR on O-HAZE and 0.05 in SSIM on NH-HAZE compared to the second-best methods, indicating a significant boost in reconstruction fidelity and structural accuracy.
Also, on Dense-HAZE, our method achieves the highest SSIM score while maintaining a comparable PSNR, whereas DAD falls short in SSIM.
These results indicates that our approach not only outperforms existing methods but also produces superior perceptual quality with fewer artifacts and sharper details.
\begin{table*}[t]
    \centering\resizebox{0.8\textwidth}{!}{
    \begin{tabular}{l|cc|cc|cc|cc}
    \toprule
    \multirow{2}{*}{Method}&\multicolumn{2}{c}{NH-HAZE~\cite{ancuti2020nh}}& \multicolumn{2}{c}{Dense-HAZE~\cite{ancuti2019dense}}& \multicolumn{2}{c}{I-HAZE~\cite{ancuti2018ihaze}}& \multicolumn{2}{c}{O-HAZE~\cite{ancuti2018ohaze}}\\
    \cmidrule(r){2-3} \cmidrule(r){4-5} \cmidrule(r){6-7} \cmidrule(r){8-9}
    & PSNR$\uparrow$ & SSIM$\uparrow$& PSNR$\uparrow$ & SSIM$\uparrow$& PSNR$\uparrow$ & SSIM$\uparrow$& PSNR$\uparrow$ & SSIM$\uparrow$ \\
    \midrule
    DAD~\cite{shao2020dad} & 14.34& 0.56 &\textbf{13.51}&0.46&18.02&0.80 &18.36&0.75\\
    PSD~\cite{chen2021psd} & 10.62& 0.52 &9.74&0.43&13.79&0.74&11.66&0.68\\
    D4~\cite{yang2022self} & 12.67& 0.50&11.50&0.45&15.64&0.73&16.96&0.72 \\
    RIDCP~\cite{wu2023ridcp}& 12.32& 0.53 & 9.85&0.45&16.88&0.78&16.52&0.72 \\
    CORUN~\cite{fang2024real}&  11.87&0.56 & 9.47&0.52&       17.14&      0.83&      18.20& 0.83      \\
    \midrule
    \midrule
    HazeFlow&\textbf{14.49}&\textbf{0.61}&11.39&\textbf{0.56} &\textbf{18.37}&\textbf{0.83}&\textbf{18.79}&\textbf{0.84}\\
    \bottomrule
    \end{tabular}}
    \caption{Quantitative results on paired dataset (NH-HAZE~\cite{ancuti2020nh}, Dense-HAZE~\cite{ancuti2019dense}, I-HAZE~\cite{ancuti2018ihaze},O-HAZE~\cite{ancuti2018ohaze}). Best results are \textbf{bolded}.}
    \label{tab:supp_paired}
\end{table*}

\section{Additional Visual Results with SOTA}
\label{sup:additional_results}
\begin{figure*}
    \centering
    \includegraphics[width=\linewidth]{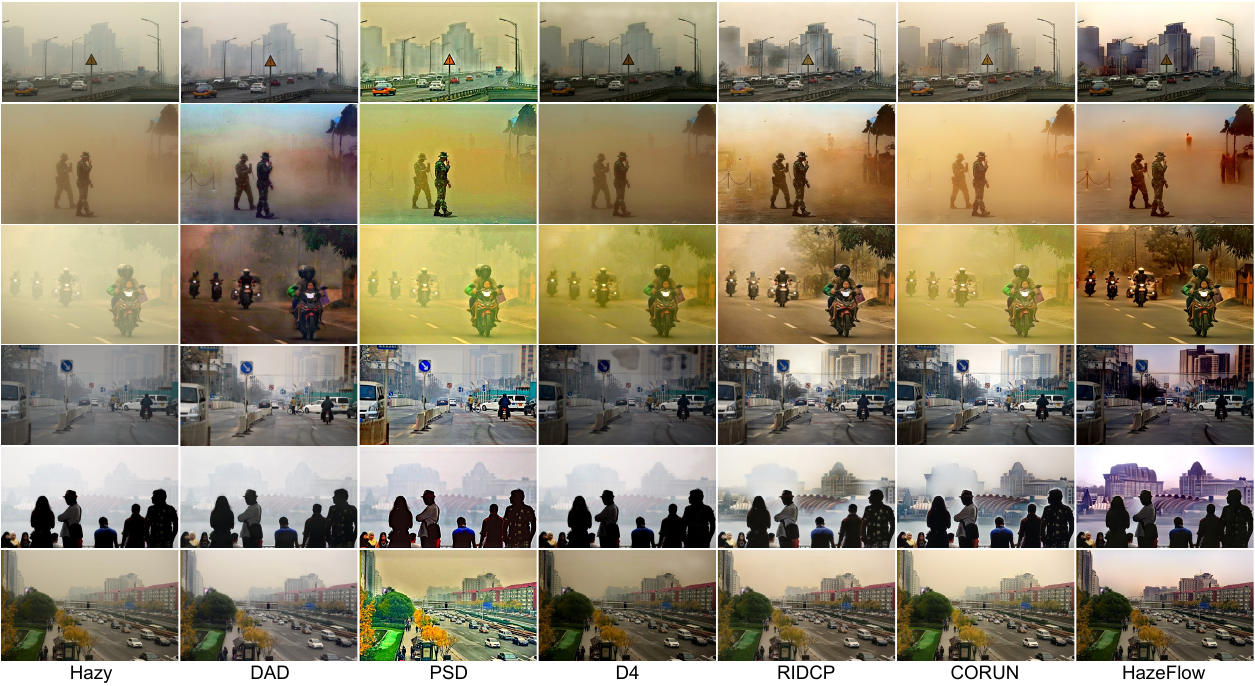}
    \caption{Additional visual comparison on RTTS dataset~\cite{li2018benchmarking}}
    \label{fig:supp_sota_rtts}
\end{figure*}

\begin{figure*}
    \centering
    \includegraphics[width=\linewidth]{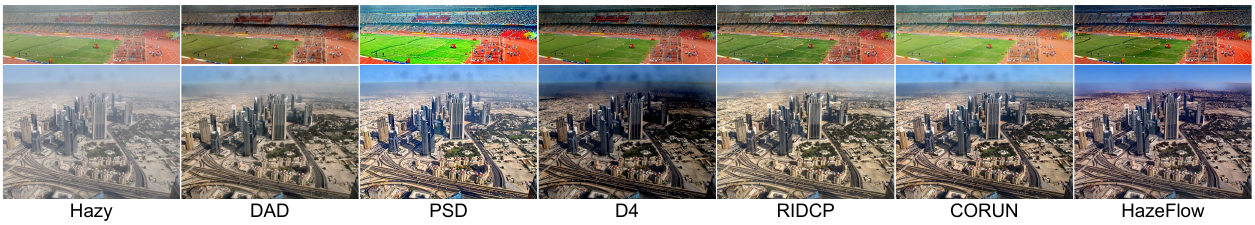}
    \caption{Additional visual comparison on Fattal's dataset~\cite{fattal2008single}}
    \label{fig:supp_sota_fattal}
\end{figure*}

\begin{figure*}
    \centering
    \includegraphics[width=\linewidth]{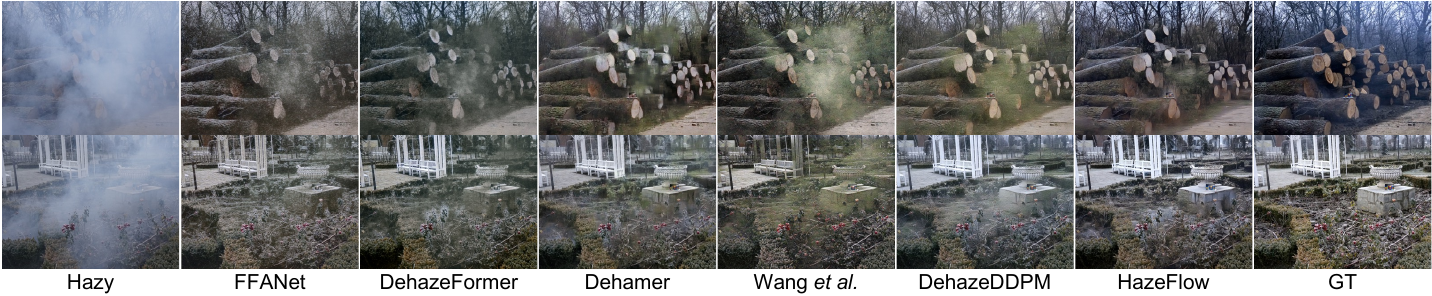}
    \caption{Additional visual comparison on NH-HAZE dataset~\cite{ancuti2020nh}.}
    \label{fig:supp_sota_NH}
\end{figure*}

\begin{figure*}
    \centering
    \includegraphics[width=\linewidth]{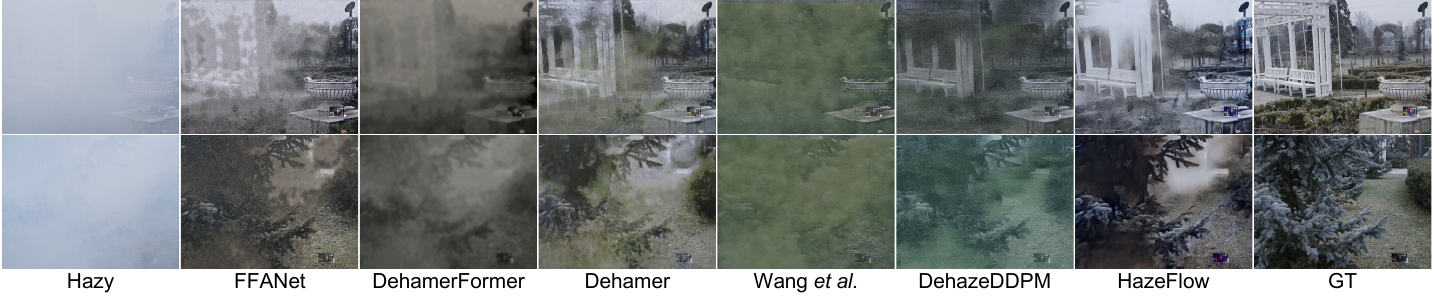}
    \caption{Additional visual comparison on Dense-HAZE dataset~\cite{ancuti2019dense}.}
    \label{fig:supp_sota_Dense}
\end{figure*}

For a comprehensive comparison with other models, we provide additional visual comparison on RTTS~\cite{li2018benchmarking} in Fig.~\ref{fig:supp_sota_rtts} and Fattal's dataset~\cite{fattal2008single}
in Fig.~\ref{fig:supp_sota_fattal}.
These figures demonstrate that HazeFlow removes haze more effectively and reduces artifacts better compared to other models.
Notably, HazeFlow uniquely preserves structural details in heavily obscured distant regions (e.g., buildings, trees), where competing methods fail to recover fine edges.  
Furthermore, we provide additional comparisons for paired datasets.
Visual results for NH-HAZE can be found in Fig.\ref{fig:supp_sota_NH}, and Dense-HAZE in Fig.\ref{fig:supp_sota_Dense}.
These results highlight that HazeFlow removes deeper haze and significantly reduces artifacts compared to other models.

  




